\documentclass[10pt, twocolumn]{article}

\usepackage[utf8]{inputenc} 
\usepackage[T1]{fontenc}    
\usepackage{hyperref}       
\usepackage{url}            
\usepackage{booktabs}       
\usepackage{amsfonts}       
\usepackage{nicefrac}       
\usepackage{microtype}      
\usepackage{xcolor}         
\usepackage{subfigure}
\usepackage{stackengine}
\usepackage{multirow}
\usepackage{multicol}
\usepackage{makecell}
\usepackage{pifont}

\usepackage{amsmath}
\usepackage{amssymb}
\usepackage{mathtools}
\usepackage{amsthm}
\usepackage{cleveref}
\theoremstyle{plain}
\newtheorem{theorem}{Theorem}

 \newtheorem{lemma}{Lemma}
 
 \crefname{assumption}{assumption}{assumptions}
 \Crefname{assumption}{Assumption}{Assumptions}

\usepackage{hyperref}

\usepackage[round]{natbib}

\allowdisplaybreaks[4]

\date{}

\title{Rethinking PGD Attack: Is Sign Function Necessary?}

\author{Junjie Yang$^1$, Tianlong Chen$^2$, Xuxi Chen$^3$, Zhangyang Wang$^3$, Yingbin Liang$^1$\\
$^1$The Ohio State University, $^2$MIT@CSAIL, $^3$University of Texas at Austin }

\begin{document}

\maketitle

\begin{abstract}
Neural networks have demonstrated success in various domains, yet their performance can be significantly degraded by even a small input perturbation. Consequently, the construction of such perturbations, known as adversarial attacks, has gained significant attention, many of which fall within "white-box" scenarios where we have full access to the neural network. Existing attack algorithms, such as the projected gradient descent (PGD), commonly take the \textit{sign} function on the raw gradient before updating adversarial inputs, thereby neglecting gradient magnitude information. In this paper, we present a theoretical analysis of how such \textit{sign}-based update algorithm influences step-wise attack performance, as well as its caveat. We also interpret why previous attempts of directly using raw gradients failed. Based on that, we further propose a new \textit{raw gradient descent} (RGD) algorithm that eliminates the use of \textit{sign}. Specifically, we convert the constrained optimization problem into an unconstrained one, by introducing a new hidden variable of non-clipped perturbation that can move beyond the constraint. The effectiveness of the proposed RGD algorithm has been demonstrated extensively in experiments, outperforming PGD and other competitors in various settings, without incurring any additional computational overhead. The codes is available in \href{https://github.com/JunjieYang97/RGD}{https://github.com/JunjieYang97/RGD}.


\end{abstract}

\section{Introduction}
Neural network has been widely adopted in many areas, e.g., computer vision~\citep{NIPS2012_c399862d, he2016deep} and natural language processing~\citep{hochreiter1997long}. Generally, a well-trained neural network can make very accurate prediction when classifying image classes. However, existing works~\citep{goodfellow2014explaining, madry2018towards,zhang2020geometry,zhang2021causaladv} have shown that merely tiny perturbations of the neural network input, which would not affect human judgment, might cause significant mistakes for the network output. These perturbations can be intentionally generated using various algorithms, usually referred to as adversarial attacks. Adversarial attacks are commonly categorized into white-box attacks and black-box attacks. In white-box attacks, we have access to all neural network parameters, allowing us to exploit this information (e.g., through back-propagation) to generate adversarial inputs. In contrast, the neural network architectures are inaccessible in the scenario of black-box attacks, and one can only test different inputs and their corresponding outputs to identify successful adversarial examples.

This work focuses exclusively on the white-box scenario, particularly for the $L_\infty$ norm-based projected gradient descent (PGD)~\citep{madry2018towards} attack. In $L_\infty$ norm attacks, the learned perturbation $\delta$ is constrained within an $\epsilon$-ball, ensuring that the absolute magnitudes of all pixel values do not exceed $\epsilon$. Under these constraints, most white-box algorithms employ the "signed gradient" to maximize the perturbation values and approach the $\epsilon$ boundary, thereby generating stronger attacks. For instance, the Fast Gradient Sign Method (FGSM)~\citep{goodfellow2014explaining} initially introduces the sign operation for one-step perturbation update. Subsequently, \citet{madry2018towards} proposes the PGD to iteratively update the perturbation using the signed gradient. Following-up PGD-based approaches, such as MIM~\citep{dong2018boosting} and Auto-Attack~\citep{croce2020reliable}, unanimously follow the practice of using the sign function. Previous work~\citep{agarwal2020role} has empirically demonstrated that the "signed gradient" significantly outperforms the raw gradient in FGSM context. However, the signed gradient, compared to the raw gradient, abandons magnitude information in PGD, which may compromise the adversary information too. This motivates the following questions:


\begin{list}{$\bullet$}{\topsep=0.01in \leftmargin=0.2in \rightmargin=0.1in \itemsep =0.01in}
	\item 
	Despite the loss of gradient value information, sign-based attacks remain the preferred choice in $L_{\infty}$ norm scenarios. What are the critical factors in determining the quality of perturbations in PGD? Why does the raw gradient fail to generate more effective adversarial attacks?
\end{list}


Note that the raw gradient is found effective in $L_2$ contexts. For instance, the C\&W~\citep{carlini2017towards} attack, a widely used algorithm, utilizes the raw gradient and incorporates constraint terms and variable changes for adversarial attacks. Therefore, more pertinent questions arise:
\begin{list}{$\bullet$}{\topsep=0.01in \leftmargin=0.2in \rightmargin=0.1in \itemsep =0.01in}
	\item Why does the "signed gradient" appear necessary in $L_\infty$ attacks? What are the pros and cons of the sign function?  Can we achieve more general attack success (including $L_\infty$) without the ``sign" operation, and can that even outperform signed gradients?
\end{list}

\subsection{Main Contributions}
This work rigorously analyzes how the perturbation update method affects the adversarial quality and further empirically shows why raw gradient is not favored in the current PGD algorithm. Moreover, we propose a new raw gradient-based algorithm that outperforms vanilla PGD across various scenarios without introducing any extra computational cost.


In this study, our first objective is to theoretically characterize the effect of update mechanisms on adversarial samples at each step. We observe that a stronger attack in the previous step, or a larger perturbation change, leads to a greater attack improvement in the subsequent step. Furthermore, our empirical observations reveal that in the later update steps, vanilla PGD, which employs the sign function, exhibits a larger change in perturbation compared to raw gradient-based PGD. Such a larger change facilitates more pixels approaching the $\epsilon$ boundary, indicating a stronger attack update based on the theoretical results. This explains the preference for using the signed gradient in PGD.


Next, we point out that the failure of the raw gradient in PGD is not only due to the insufficient update on its own,  but also attributed to the clipping design. In PGD, all perturbations are clipped within the $\epsilon$ ball per step. This design results in the loss of significant magnitude information for the raw gradient update, particularly when most pixels approach the $\epsilon$ boundary. Instead, we introduce a new hidden variable of non-clipped perturbation for updating. In this way, the $L_\infty$ norm-based adversarial attack, a constrained optimization problem, is transformed into an unconstrained optimization problem, where the proposed hidden variable is allowed to surpass the $\epsilon$ boundary. Extensive experiments further demonstrate that our proposed method significantly improves the performance of raw update-based PGD and even outperforms vanilla PGD, without incurring additional computational overhead. 

\section{Related Works}
\label{sec:related-work} 
Adversarial robustness has been explored in recent years. \citet{szegedy2013intriguing} were the first to mention that imperceptible perturbations can cause networks to misclassify images. Then, based on gradients with respect to the input, \citet{goodfellow2014explaining} proposed the one-step FGSM for generating adversarial examples. Furthermore, iterative-based adversarial attacks such as BIM~\citep{kurakin2018adversarial} and PGD~\citep{madry2018towards} have been proposed for white-box attacks. The difference between these algorithms lies in whether they adopt random initialization or not.
The C\&W attack~\citep{carlini2017towards} introduced a regularized loss parameterized by $c$ and demonstrated that defensive distillation does not significantly enhance robustness. DeepFool~\citep{moosavi2016deepfool} generated perturbations by projecting the data onto the closest hyperplane.
Auto-Attack~\citep{croce2020reliable} combines four different attacks, two of which are PGD variants.
Regarding the role of the "sign" function, \citet{agarwal2020role} demonstrated that gradient magnitude alone cannot result in a successful FGSM attack. Meanwhile, by manipulating the images in the opposite direction of the gradient, the classification error rates can be significantly reduced. Furthermore, there are several works~\citep{zhang2022revisiting, liu2019signsgd, shaham2015understanding} to study the sign function within the bilevel optimization or adversarial training frameworks.

Black-box attacks have also received significant attention in the field. For instance, \citet{chen2017zoo} introduced the use of zeroth-order information to estimate the neural network gradients in order to generate adversarial examples. In a different approach, \citet{ilyas2018black} incorporated natural evolutionary strategies to enhance query efficiency. Furthermore, \citet{al2020sign} proposed a sign-based gradient estimation approach, which replaced continuous gradient estimation with binary black-box optimization.
The Square Attack algorithm, proposed by \citet{andriushchenko2020square}, introduced randomized localized square-shaped updates, resulting in a substantial improvement in query efficiency. Another approach, Rays~\citep{chen2020rays}, transformed the continuous problem of finding the closest decision boundary into a discrete problem and achieved success in hard-label contexts.

\section{Methodology}
\label{sec:methodology}
In this section, we introduce the formulation of adversarial attacks and present our algorithm. Let us consider the input $x \in \mathbb{R}^{n}$ with its true label $y$. We define the prediction model $f(x; w)$, where $w$ represents the model's parameters. The loss function, denoted as $l(f(x; w), y)$, quantifies the discrepancy between the model's output $f(x; w)$ and the true label $y$.
Hence, our objective in $L_\infty$ norm based adversarial attack is to maximize the loss function $g(x + \delta)$, while constraining the perturbation $\delta$ within an $\epsilon$-ball, as follows:
\begin{align*}
\max_{\delta:\|\delta\|_\infty\leq \epsilon} l(f(x+\delta);w),y) = g(x+\delta).
\end{align*}
In practice, we adopt 
 $\epsilon$-ball projection $\sigma_\epsilon(\cdot)$ to construct the projected perturbation $\delta_c = \sigma_\epsilon(\delta)$ and it satisfies $\|\delta_c\|_\infty \leq \epsilon$. The $\sigma_\epsilon(\cdot)$ operation can be characterized as follows:
\begin{align*}
    \delta_c = \sigma_\epsilon(\delta)=\max(\min(\delta, \epsilon), -\epsilon).
\end{align*}
Note that $\sigma_\epsilon$ projection operates element-wisely.

Inspired by Fast Gradient Sign Method (FGSM)~\citep{goodfellow2014explaining}, most attack algorithms utilize the sign function for perturbation update where Projected Gradient Descent (PGD)~\citep{madry2018towards} is widely studied and applied. Specifically, The perturbation $\delta$ is iteratively updated using the signed gradient in PGD.
If we represent the perturbation at the $t$-th step as $\delta^t$ $(t=0, \ldots, T-1)$, the update procedure in PGD can be described as follows:
\begin{align}
\label{eq:vanilla_pgd_update}
    \textit{(Vanilla PGD)} \quad 
    \delta^{t+1} = \delta^t_c+\alpha \text{sign}(\nabla_\delta g(x+\delta^t_c)),
\end{align}
where $\delta_c^t=\sigma_\epsilon(\delta^t)$ represents the element-wise clipped (or projected) perturbation at the $t$-th step. The \textit{sign()} function assigns a value of 1 to positive elements and -1 to negative elements. The adversarial performance is evaluated by measuring the performance of the clipped final step perturbation, denoted as $g(x+\delta_c^T)$.

To maximize the adversarial loss $g(x+\delta_c^T)$, a natural question arises: can we eliminate the \textit{sign()} function in eq.\ref{eq:vanilla_pgd_update}, considering that the raw gradient contains more information before being projected into a binary value? A naive variant of PGD with raw updates is defined:
\begin{align}
\label{eq:raw_pgd_update}
    \textit{(PGD raw gradient)} \quad \delta^{t+1} = \delta^t_c+\alpha \nabla_\delta g(x+\delta^t_c),
\end{align}
where we simply remove the \textit{sign()} function. However, our experimental results have indicated that update in eq.\ref{eq:raw_pgd_update} result in worse performance compared to the signed gradient update in terms of robust accuracy.

The failure of utilizing raw gradient arises from the usage of clipped perturbation for update. In \Cref{subsec:boundary_study}, we show that during the update process, more than half of the perturbation elements cross and are projected into the $\epsilon$ boundary. Consequently, a significant amount of \textit{raw gradient magnitude} information is eliminated in $\delta_c^t$, making its adoption for update ineffective in strengthening the attack. 

To address this issue, we introduce a hidden unclipped perturbation $\delta^t$ for update and propose \textbf{Raw Gradient Descent (RGD)} algorithm, outlined as follows:
\begin{align}
  \textit{(Proposed RGD)}\quad  \delta^{t+1}=\delta^t + \alpha \nabla_\delta g(x+\delta_c^t).
\end{align}
It is important to note that the intermediate update of $\delta^{t+1}$ depends on the unclipped $\delta^t$ and is allowed to cross the $\epsilon$ boundary. The projection restriction is solely applied to the raw gradient $\nabla_\delta g(x+\delta_c^t)$. As a result, the constrained optimization problem regarding $\delta_c^t$ is transformed into an unconstrained optimization problem concerning $\delta^t$, with the restriction implicitly applied in the gradient computation $\nabla_\delta g(x+\delta_c^t)$. The last-iteration output $\delta^{T}$ will be clipped to fit the constraint, as the final output. This design allows the intermediate perturbation to learn the genuine adversarial distribution without any loss of magnitude information. Its effectiveness has been demonstrated through experiments in \Cref{sec:experiment}, where the clipped objective function $g(x+\delta_c^T)$ was utilized. Furthermore, our proposed algorithm does not incur any additional computational cost compared to the vanilla PGD in \cref{eq:vanilla_pgd_update}, except saving the extra term $\delta^t$.


\begin{table*}
    \centering
    \begin{tabular}{c|c|c|c|c|c|c|c|c}
        \Xhline{1.5pt}
        & Algorithm & 1 & 2 & 3 & 4 & 5 & 6 & 7 \\
        \hline
        \multirow{3}{*}{\makecell{Robust \\Accuracy (\%)}} & PGD & 65.51 & 52.8 & 46.74 & 44.89 & 43.84 & 43.43 & 43.19 \\
        \cline{2-9}
        &PGD (raw)  & 58.74 & 50.36 & 46.7 & 45.58 & 44.98 & 44.68 & 44.54  \\
        \cline{2-9}
        & RGD  & 55.65 & 47.58 & 45.14 & 43.97 & 43.4 & 43.04 & \textbf{42.88}\\
        \hline
        \hline
        \multirow{3}{*}{\makecell{Boundary\\ Ratio (\%)}} & PGD & 31.7 & 52.5 & 74.5 & 78.1 & 83.2 & 85.0 & 86.9\\
        \cline{2-9}
        &PGD (raw) & 24.2 & 39.6 & 47.8 & 52.2 & 55.3 & 57.4 & 59.1 \\
        \cline{2-9}
        &RGD & 34.5 & 52.5 & 60.1 & 64.3 & 67.0 & 69.0 & 70.6\\
        \Xhline{1.5pt}
    \end{tabular}
    \caption{Comparison of update algorithms for robust accuracy and perturbation pixel-wise boundary ratio in different steps.}
    \label{tab:pertu_illu}
\end{table*}

\section{Understanding How Update Influences PGD Performance}
In this section, we begin by conducting a theoretical analysis of the PGD update and examine how the step-wise update influences the  attack performance. Subsequently, we integrate our theoretical findings with experimental results to elucidate why the sign operation has been favored in PGD.

\begin{figure*}[ht]
    \centering
    PGD Coefficient Histogram\\
    \subfigure{\includegraphics[width=0.26\linewidth]{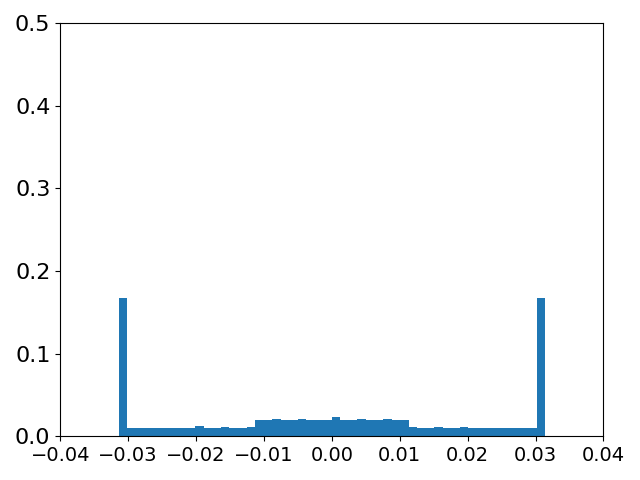}}
    \hspace{-18pt}
    \subfigure{\includegraphics[width=0.26\linewidth]{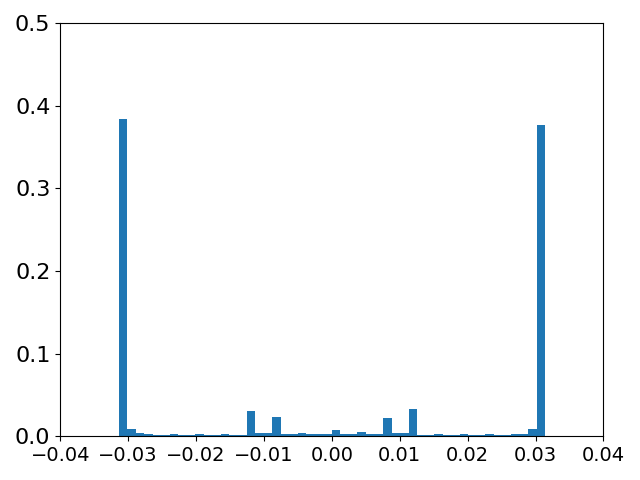}}
    \hspace{-18pt}
    \subfigure{\includegraphics[width=0.26\linewidth]{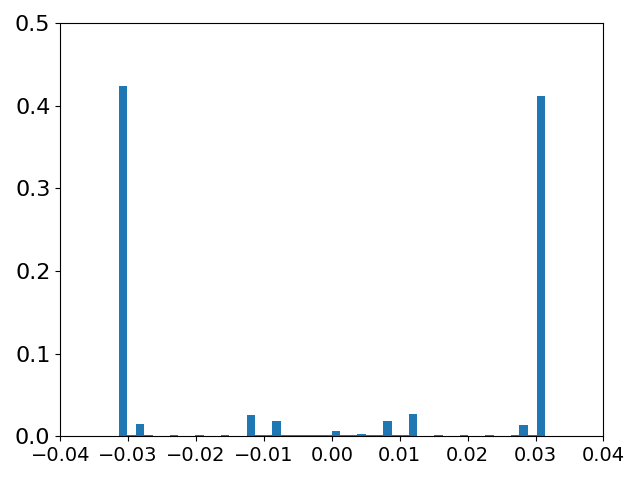}}
    \hspace{-18pt}
    \subfigure{\includegraphics[width=0.26\linewidth]{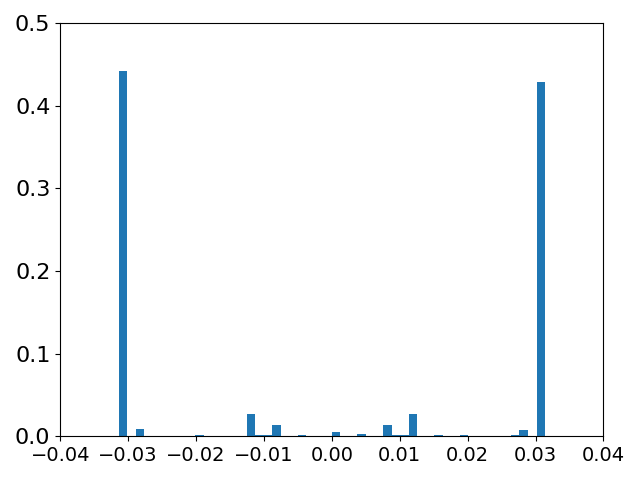}}
    \\
    PGD (raw) Coefficient Histogram\\
    \subfigure{\includegraphics[width=0.26\linewidth]{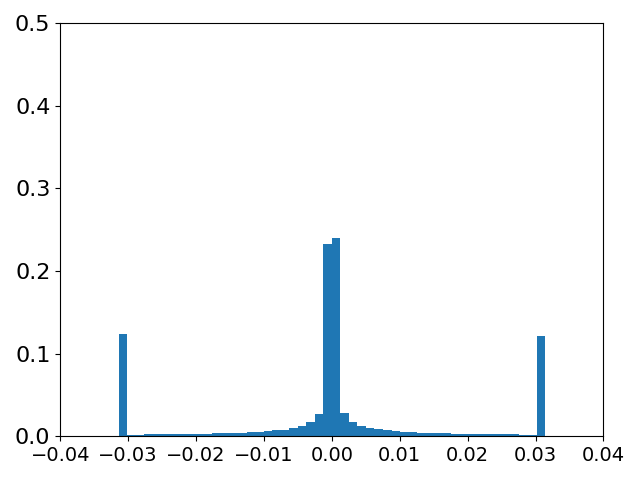}}
    \hspace{-18pt}
    \subfigure{\includegraphics[width=0.26\linewidth]{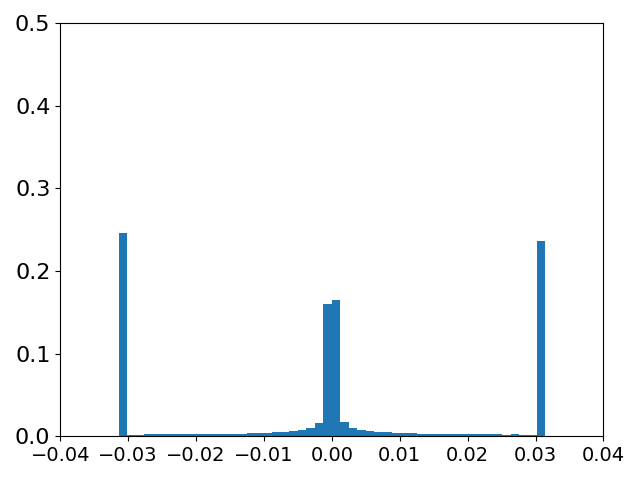}}
    \hspace{-18pt}
    \subfigure{\includegraphics[width=0.26\linewidth]{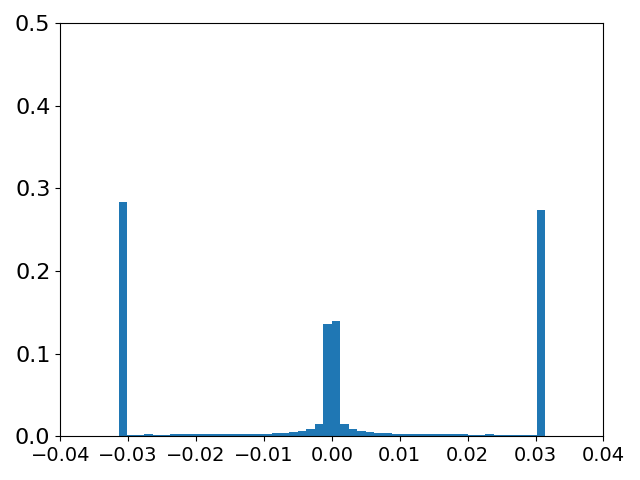}}
    \hspace{-18pt}
    \subfigure{\includegraphics[width=0.26\linewidth]{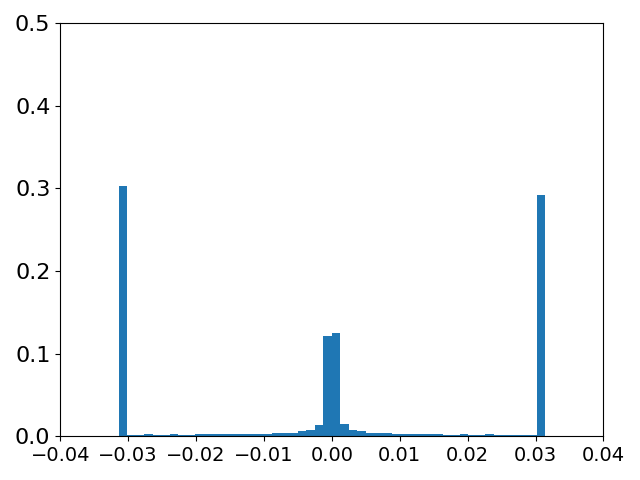}}
    \\
    RGD Coefficient Histogram \\
    \stackunder[5pt]{\includegraphics[width=0.26\linewidth]{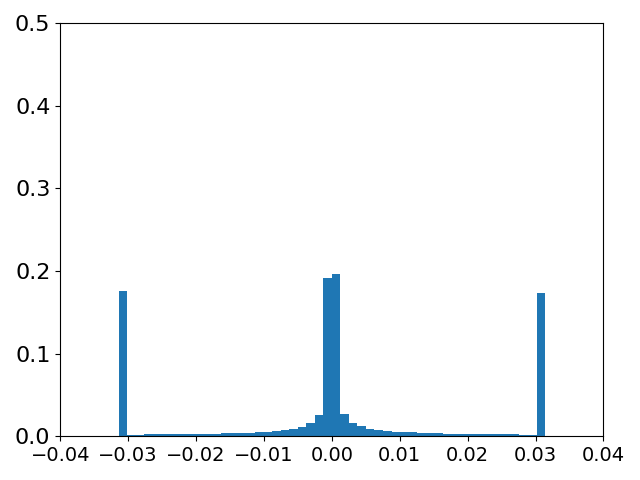}}{step1}
    \hspace{-18pt}
    \stackunder[5pt]{\includegraphics[width=0.26\linewidth]{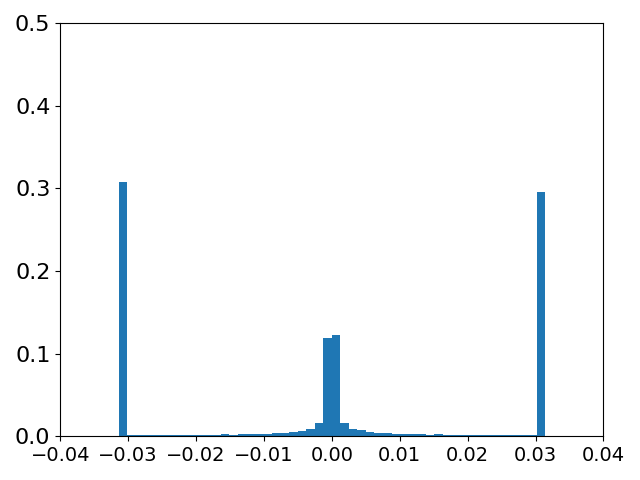}}{step3}
    \hspace{-18pt}
    \stackunder[5pt]{\includegraphics[width=0.26\linewidth]{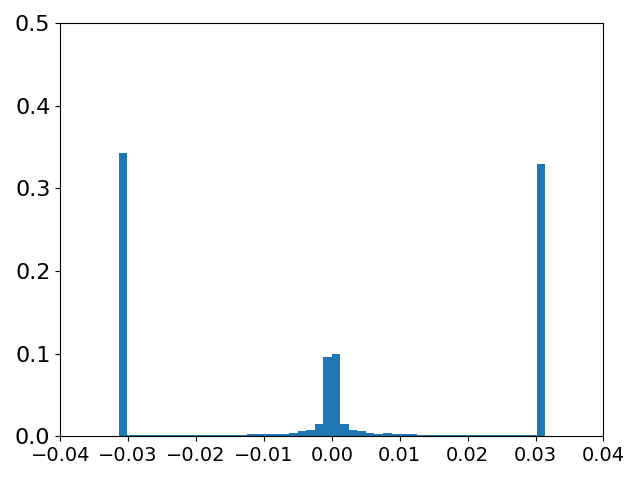}}{step5}
    \hspace{-18pt}
    \stackunder[5pt]{\includegraphics[width=0.26\linewidth]{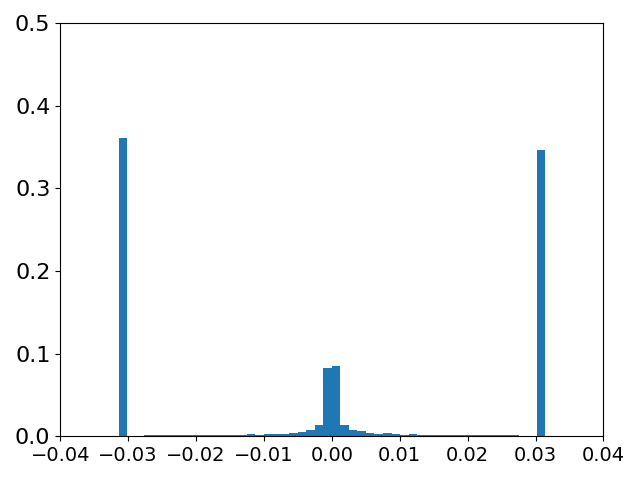}}{step7}
    \caption{Comparison of update algorithms for perturbation pixel distribution in different steps.}
    \label{fig:pertu_illu}
\end{figure*}


\subsection{Theoretical Insights}
\label{subsec:theoy_insight}
In this part, we characterize the influence of the update procedure on the adversarial gain $g(x+\delta_c^{t+1})-g(x+\delta_c^t)$ at each step $t$. Inspired by existing works~\citep{dugradient, arora2019fine}, we consider a model function with one ReLU~\citep{nair2010rectified} hidden layer, defined as follows:
\begin{align}
\label{eq:relu_formulation}
    f(x;w) = w_1^Th(w_2*x),
\end{align}
where $h(x)=\max(0,x)$ denotes the element-wisely ReLU function, $x\in\mathbb{R}^{n\times 1}$ denotes the input, $ w_1\in\mathbb{R}^{m\times 1}$ and $ w_2\in\mathbb{R}^{m\times n}$ denote output weight and first-layer vector weight respectively. Assuming the Mean Square Error (MSE) loss, the loss function $g(x+\delta_c)$ is defined as follows:
\begin{align}
\label{eq:mse_loss}
    g(x+\delta_c) = \frac{1}{2}[f(x+\delta_c;w)-y^\ast(x)]^2,
\end{align}
where $y^\ast(x)$ denotes the real label value for input $x$.

\begin{theorem}
Considering $g(x)=\frac{1}{2}[w_1^Th(w_2*x)-y^\ast(x)]^2$, activation function $h(x)$ as ReLU, we define $|\cdot| $ as element wise absolute operation, then the adversarial step gain $g(x+\delta_c^{t+1})-g(x+\delta_c^t)$ is bounded as follows:
\label{thm:main}
\begin{align*}
    g(x+\delta_c^{t+1})- g(x+&\delta_c^t) \leq  \frac{\sqrt{2}}{2} |w_1^T||w_2||\delta_c^{t+1}-\delta_c^{t}| \\
    & \left(\sqrt{g(x+\delta_c^{t+1})} + \sqrt{g(x+\delta_c^{t})}\right),
\end{align*}
where $\delta_c^t$ denotes the clipped perturbation in $t$-th step.
\end{theorem}

The theorem demonstrates that the adversarial gain in step $t$ depends on two factors: (i) The magnitude of perturbation change, $\delta_c^{t+1}-\delta_c^t$. If the update algorithm results in a large perturbation change, the adversarial gain will be increased. (ii) The previous adversarial loss $g(x+\delta_c^t)$, and the updated loss $g(x+\delta_c^{t+1})$. It is important to note that the adversarial loss is non-negative, and we expect the updated $g(x+\delta_c^{t+1})$ to be greater than the previous $g(x+\delta_c^t)$. Therefore, the step gain $g(x+\delta_c^{t+1})-g(x+\delta_c^t)$ is highly influenced by the previous loss $g(x+\delta_c^t)$. As a result, a larger previous adversarial loss $g(x+\delta_c^t)$ implies a better step gain.
In summary, if our update algorithm induces a significant perturbation change $\delta_c^{t+1}-\delta_c^t$ and starts with a substantial initial adversarial loss $g(x+\delta_c^t)$, it will lead to a larger adversarial gain $g(x+\delta_c^{t+1})-g(x+\delta_c^t)$.

\begin{table*}
    \centering
    \begin{tabular}{c|c|c|c|c}
     \Xhline{1.5pt}
     Dataset & Method & PGD &
     PGD(raw) & RGD \\ 
     \hline
      \multirow{4}{*}{\makecell{CIFAR-10 \\ ($\epsilon=8/255$)}} & WRN-28\citep{dingmma} & 52.64\tiny $\pm 0.052 $ & 53.98\tiny $\pm 0.067$ & \textbf{52.56}\tiny $\pm 0.013$ \\ \cline{2-5}
     & PreRN-18\citep{wongfast} &47.45\tiny $\pm 0.052$ & 47.50\tiny $\pm 0.013$ & \textbf{47.31}\tiny $\pm 0.005$ \\ \cline{2-5}
     & RN-18\citep{addepalli2022efficient} & 57.06\tiny $\pm 0.044$ & 57.11\tiny $\pm 0.005$ & \textbf{57.03}\tiny $\pm 0.000$ \\ \cline{2-5}
     & RN-18\citep{robustness} & 43.23\tiny $\pm 0.023$ & 44.50\tiny $\pm 0.015$ & \textbf{42.87}\tiny $\pm 0.010$ \\
     \hline
     \multirow{4}{*}{ \makecell{CIFAR-10\\ ($\epsilon=16/255$)}} & WRN-28\citep{dingmma} & 37.53\tiny $\pm 0.143$ & 40.93\tiny $\pm 0.068$ & \textbf{37.19}\tiny $\pm 0.024$ \\
     \cline{2-5}
     & PreRN-18\citep{wongfast} &13.32\tiny $\pm 0.073$ & 14.98\tiny $\pm 0.012$ & \textbf{12.99}\tiny $\pm 0.015$ \\
     \cline{2-5}
     & RN-18\citep{addepalli2022efficient} & 25.57\tiny $\pm 0.094$ & 25.87\tiny $\pm 0.016$ & \textbf{25.27}\tiny $\pm 0.006$ \\
     \cline{2-5}
     & RN-18\citep{robustness} & 14.19\tiny $\pm 0.033$ & 19.11\tiny $\pm 0.048$ & \textbf{13.37}\tiny $\pm 0.028$ \\
     \hline
     \hline
     \multirow{5}{*}{\makecell{CIFAR-100\\ ($\epsilon=16/255$)}} & WRN-28\citep{wang2023better} & 21.10\tiny $\pm 0.075$ & 20.72\tiny $\pm 0.000$ & \textbf{20.61}\tiny $\pm 0.000$ \\ \cline{2-5}
     & XCiT\citep{debenedetti2022light} & 15.28\tiny $\pm 0.067$ & 15.44\tiny $\pm 0.000$ & \textbf{15.28}\tiny $\pm 0.004$ \\ \cline{2-5}
     & WRN-34\citep{addepalli2022efficient} & 16.37\tiny $\pm 0.010$ & 16.16\tiny $\pm 0.019$ & \textbf{15.94}\tiny $\pm 0.013$ \\ \cline{2-5}
     & RN-18\citep{addepalli2022efficient} & 14.45\tiny $\pm 0.082$ & 14.40\tiny $\pm 0.008$ & \textbf{14.20}\tiny $\pm 0.010$ \\ \cline{2-5}
     & PreRN-18\citep{rice2020overfitting} & 6.14\tiny $\pm 0.035$ & 7.36\tiny $\pm 0.018$ & \textbf{5.78}\tiny $\pm 0.010$ \\
     \hline
     \hline
     \multirow{3}{*}{\makecell{ImageNet\\ ($\epsilon=4/255$)}} & WRN-50\citep{salman2020adversarially} & 42.02\tiny $\pm 0.041$& 42.57\tiny $\pm 0.032$ & \textbf{41.88}\tiny $\pm 0.015$ \\
     \cline{2-5}
     & RN-50\citep{wongfast} & 29.0\tiny $\pm 0.028$ & 28.56\tiny $\pm 0.013$ & \textbf{27.85}\tiny $\pm 0.027$ \\
     \cline{2-5}
     & RN-18\citep{salman2020adversarially} & 30.08\tiny $\pm 0.054$ & 30.45\tiny $\pm 0.010$ & \textbf{30.08}\tiny $\pm 0.008$ \\
     \hline
     \multirow{3}{*}{\makecell{ImageNet\\ ($\epsilon=8/255$)}} & WRN-50\citep{salman2020adversarially}& 19.35\tiny $\pm 0.041$ & 21.75\tiny $\pm 0.050$ & \textbf{18.97}\tiny $\pm 0.032$ \\
     \cline{2-5}
     & RN-50\citep{wongfast} & 11.77\tiny $\pm 0.030$ & 12.31\tiny $\pm 0.016$ & \textbf{11.03}\tiny $\pm 0.010$ \\
     \cline{2-5}
     & RN-18\citep{salman2020adversarially} & 12.88\tiny $\pm 0.027$ & 13.69\tiny $\pm 0.010$ & \textbf{12.68}\tiny $\pm 0.008$ \\
     \Xhline{1.5pt}
\end{tabular}
    \caption{Robust accuracy comparison of PGD, PGD with raw update and RGD for 7-step attack. The methods are abbreviated. XCiT-S12:XCiT, WideResNet:WRN, ResNet:RN, and PreActResNet:PreRN.}
    \label{tab:pgd_attack}
\end{table*}

\subsection{Empirical Study}
\label{subsec:boundary_study}
We next conduct a further study to analyze how the update process shapes adversarial samples in practice. Specifically, we target the robust PGD model~\footnote{The robust model used can be found at \url{https://github.com/ndb796/Pytorch-Adversarial-Training-CIFAR} with CIFAR10 $\epsilon=8/255$ $L_\infty$ setting.} on the CIFAR10 testing dataset within the $\epsilon=8/255$ ball. The perturbation distribution at each step for three update algorithms (PGD, PGD with raw update, proposed RGD) is illustrated in \Cref{fig:pertu_illu}. The results reveal that, within the limited 7 steps, the majority of pixels (86.9\%) converge to the boundary in the PGD update, whereas its corresponding raw update shows a lower boundary ratio (59.1\%) with a considerable number of pixels remaining stuck at the zero initial point. 
More detailed information regarding the boundary ratio and robust accuracy can be found in \Cref{tab:pertu_illu}. It reveals that PGD achieves a lower robust accuracy ($43.19\%$) compared to its raw update version ($44.54\%$). Considering the theoretical results presented in \Cref{thm:main}, which indicate that larger perturbation changes lead to better performance gains, we can conclude that the success of PGD, in contrast to PGD with naive raw update, can be attributed to its "sign" ability, which facilitates more pixel changes and convergence to the boundary.

Furthermore, it is worth noting that RGD outperforms PGD in terms of attack performance, despite having fewer boundary pixels (70.6\% v.s. 86.9\%). This can be attributed to the fact that the real adversarial distribution in RGD exhibits better attack performance, as indicated by the higher values of $g(x+\delta_c^t)$, resulting in larger adversarial gains at each step, as demonstrated in \Cref{tab:pertu_illu}. Consequently, RGD is able to learn a better adversarial distribution even without a significant perturbation change. It is important to highlight that PGD (raw), despite its use of raw updates, does not effectively preserve most of the magnitude information due to the clip operation. As a result, it performs significantly worse than RGD.

To ensure a fair comparison, we carefully fine-tuned the step size $\alpha$ for all three algorithms, using grid search. Following the principle of minimal robust accuracy, we set a zero initial point for both PGD (raw) and RGD, while utilizing a random initial point for PGD. Note that the choice of initial point does not influence the final boundary ratio while detailed initial comparison is available in \Cref{app:initial}.

\section{Experimental Results}
\label{sec:experiment}
In this section, we begin by comparing the proposed RGD algorithm with PGD (sign/raw updates) for adversarial attacks on different architectures, with varying adversarial perturbation levels and datasets. Then we compare RGD and PGD over adversarial training setting and showcase the robust accuracy performance boosts brought by RGD. Moreover, we utilize RGD for transfer attacks and conduct extensive experiments to validate its remarkable improvement in adversarial transferability. All experiments are conducted with \textbf{five independent runs} with different random seeds, and the standard deviations are reported to affirm results' significance. 
The step sizes $\alpha$ for all algorithms are carefully tuned through grid search. For the PGD, we grid search the step size from list (2$\epsilon$, 1.5$\epsilon$, $\epsilon$, 0.8$\epsilon$, 0.5$\epsilon$, 0.25$\epsilon$, 0.2$\epsilon$) and our RGD, PGD(raw) step size tuned from list (1, 3, 10, 30, 100, 300, 1000, 3000, 1e4, 3e4, etc). The step size for RGD and PGD(raw) is larger than PGD ones because the raw gradient over the input is small which calls for a large step size for update. Meanwhile, we utilize random initialization for PGD and zero initialization for the other methods. The initial choice is based on minimal robust accuracy principle. The detailed initial comparison is available in \Cref{app:initial}. All experiments use NVIDIA Volta V100 GPUs.

\subsection{Comparison of Algorithms for Adversarial Attack}
\label{subsec:pgd_compare}
We compare PGD with sign/raw updates and proposed RGD using a 7-step attack. 
 The datasets include CIFAR-10, CIFAR-100, and ImageNet, and we attack their respective testing or validation sets. The models attacked are sourced from RobustBench~\citep{croce2021robustbench}. The adversarial attack settings follows the approach of~\citet{ding2019advertorch}, and the results are presented in \Cref{tab:pgd_attack}.

Our results demonstrate that introduced non-clipped perturbation significantly improves the raw update method, resulting in a substantial performance boost from PGD (raw) to RGD. Furthermore, RGD outperforms PGD across various datasets, model architectures, and adversarial sizes, highlighting the superiority of RGD. It is worth noting that RGD is particularly advantageous in scenarios with larger $\epsilon$ values (e.g., $\epsilon=16/255$). This is likely because larger adversarial size allows RGD to exhibit a more effective adversarial distribution. Conversely, if a smaller $\epsilon$-ball is used, most RGD perturbation pixels will be clipped into the $\epsilon$ boundary in the final step.

\subsection{Comparison of Algorithms for Adversarial Training}

For adversarial training, we conduct experiments with different architectures and steps on CIFAR10 to compare PGD and RGD. We specifically train models including ResNet, WideResNet, and a Convolutional Neural Network, using both 5-step and 10-step attack strategies. Each model comprises 6 blocks. The robustness of these models is assessed using a 10-step Projected Gradient Descent (PGD) attack, with all experiments conducted within an $\epsilon=8/255$ constraint. Both clean and robust accuracy results are presented in \Cref{tab:at_comparison}.

\begin{table*}
	\centering
	\begin{tabular}{c|c|c|c}
		\Xhline{1.5pt}
		Setting & Method & Clean Accuracy & Robust Accuracy \\
		\hline
		\multirow{2}{*}{\makecell{ResNet \\ (Step=5)}} & PGD & 84.41 \% & 29.32 \%\\
		\cline{2-4}
		& \textbf{RGD} & 76.35\% & \textbf{44.67\% (15.35\%$\uparrow$)} \\
		\hline
		\multirow{2}{*}{\makecell{ResNet \\ (Step=10)}} & PGD & 79.41\% & 38.15 \%\\
		\cline{2-4}
		& \textbf{RGD} & 78.35\% & \textbf{47.16\% (9.01\%$\uparrow$)} \\
		\hline
		\hline
		\multirow{2}{*}{\makecell{WideResNet \\ (Step=5)}} & PGD & 89.83\% & 37.86 \%\\
		\cline{2-4}
		& \textbf{RGD} & 86.17\% & \textbf{52.15\% (14.29\%$\uparrow$)} \\
		\hline
		\multirow{2}{*}{\makecell{WideResNet \\ (Step=10)}} & PGD & 80.53\% & 48.33 \%\\
		\cline{2-4}
		& \textbf{RGD} & 86.04\% & \textbf{52.19\% (3.86\%$\uparrow$)} \\
		\hline
		\hline
		\multirow{2}{*}{\makecell{CNN \\ (Step=5)}} & PGD & 86.85\% & 26.00 \%\\
		\cline{2-4}
		& \textbf{RGD} & 82.72\% & \textbf{42.76\% (16.76\%$\uparrow$)} \\
		\hline
		\multirow{2}{*}{\makecell{CNN \\ (Step=10)}} & PGD & 82.10\% & 41.35 \%\\
		\cline{2-4}
		& \textbf{RGD} & 82.63\% & \textbf{43.03\% (1.68\%$\uparrow$)} \\
		\Xhline{1.5pt}
	\end{tabular}
	\caption{Adversarial training comparison of PGD and RGD. CNN refers to Convolutional neural network. The number in the bracket shows the robust accuracy improvements by switching PGD to RGD.}
	\label{tab:at_comparison}
\end{table*}

From the table, it can be observed that RGD significantly improves all robust accuracy across different architectures and steps. Specifically, for the setting of WideResNet step=10, RGD improves both clean and robust accuracy. For the other settings of CNN step=5, RGD sacrifices a little clean accuracy of $3\%$ but significantly improves the robust accuracy of $16\%$ than PGD. All these experimental results further validate the superiority of RGD over PGD, especially with setting of step=5.

\begin{table*}
    \centering
    \begin{tabular}{c|c|c|c|c|c}
    \Xhline{1.5pt}
      Dataset & Method & Type & PGD & PGD (raw) & RGD \\ \hline
       \multirow{7}{*}{\makecell{ImageNet\\ ($\epsilon=8/255$)}} &
        ResNet-50 & Source & 99.97\tiny $\pm 0.001$ & 96.73\tiny $\pm 0.068$ & 99.30\tiny $\pm 0.015$ \\ \cline{2-6}
        & DenseNet-121 & Clean & 54.52\tiny $\pm 0.15$ & 24.34\tiny $\pm 0.355$ & \textbf{59.42}\tiny $\pm 0.379$ \\ \cline{2-6}
        &VGG19-BN & Clean & 51.06\tiny $\pm 0.155$ & 26.51\tiny $\pm 0.241$ & \textbf{59.40}\tiny $\pm 0.358$ \\ \cline{2-6}
        &Inception-V3 & Clean & 29.04\tiny $\pm 0.349$ & 22.42\tiny $\pm 0.355$ & \textbf{34.82}\tiny $\pm 0.207$ \\ \cline{2-6}
        &PreActResNet-18 & Robust & 30.09\tiny $\pm 0.077$ & 30.51\tiny $\pm 0.118$ & \textbf{30.88}\tiny $\pm 0.027$ \\ \cline{2-6}
        &WideResNet-50 & Robust & 12.13\tiny $\pm 0.051$ & 12.55\tiny $\pm 0.07$ & \textbf{12.77}\tiny $\pm 0.079$ \\ \cline{2-6}
        &ResNet-50 & Robust & 18.14\tiny $\pm 0.096$ & 18.81\tiny $\pm 0.116$ & \textbf{19.11}\tiny $\pm 0.047$ \\ \hline \hline

      \multirow{7}{*}{\makecell{ImageNet\\ ($\epsilon=16/255$)}} & ResNet-50 & Source & 100\tiny $\pm 0.008$ & 98.08\tiny $\pm 0.194$ & 99.30\tiny $\pm 0.008$ \\ \cline{2-6}
       & DenseNet-121 & Clean & 77.58\tiny $\pm 0.221$ & 42.76\tiny $\pm 0.363$  & \textbf{81.1}\tiny $\pm 0.196$  \\
        \cline{2-6}
        &VGG19-BN & Clean & 72.67\tiny $\pm 0.142$  & 49.98\tiny $\pm 0.289$ & \textbf{80.62}\tiny $\pm 0.246$  \\
        \cline{2-6}
        &Inception-V3 & Clean & 44.04\tiny $\pm 0.414$  & 35.01\tiny $\pm 0.193$ & \textbf{54.6}\tiny $\pm 0.253$  \\
        \cline{2-6}
        &PreActResNet-18 & Robust & 34\tiny $\pm 0.041$ & 37.66\tiny $\pm 0.054$ & \textbf{38.22}\tiny $\pm 0.093$  \\
        \cline{2-6}
        &WideResNet-50 & Robust & 15.03\tiny $\pm 0.131$  & 19.12\tiny $\pm 0.113$ & \textbf{19.62}\tiny $\pm 0.078$  \\
        \cline{2-6}
        &ResNet-50 & Robust & 21.96\tiny $\pm 0.054$  & 26.81\tiny $\pm 0.183$ & \textbf{27.52}\tiny $\pm 0.109$  \\ \hline 
        \hline

        \multirow{7}{*}{\makecell{CIFAR-10\\ ($\epsilon=8/255$)}} &
       ResNet-50 & Source & 98.76\tiny $\pm 0.065$ & 78.99\tiny $\pm 0.316$ & 98.22\tiny $\pm 0.026$ \\ \cline{2-6}
        & DenseNet-121 & Clean & 65.74\tiny $\pm 0.3$ & 43.05\tiny $\pm 0.145$ & \textbf{79.66}\tiny $\pm 0.124$ \\ \cline{2-6}
        &VGG19-BN & Clean & 58.20\tiny $\pm 0.159$ & 45.87\tiny $\pm 0.341$ & \textbf{72.23}\tiny $\pm 0.155$ \\ \cline{2-6}
        &Inception-V3 & Clean & 64.28\tiny $\pm 0.17$ & 49.87\tiny $\pm 0.357$ & \textbf{77.3}\tiny $\pm 0.1$ \\ \cline{2-6}
        &WideResNet-28 & Robust & 17.41\tiny $\pm 0.039$ & 17.86\tiny $\pm 0.052$ & \textbf{18.88}\tiny $\pm 0.037$ \\ \cline{2-6}
        &{PreActResNet-18}$^1$ & Robust & 18.11\tiny $\pm 0.057$ & 18.06\tiny $\pm 0.049$ & \textbf{19.01}\tiny $\pm 0.049$ \\ \cline{2-6}
        &{PreActResNet-18}$^2$ & Robust & 21.46\tiny $\pm 0.079$ & 21.62\tiny $\pm 0.063$ & \textbf{22.54}\tiny $\pm 0.039$ \\ \cline{2-6}
       \hline \hline
       
    \multirow{7}{*}{\makecell{CIFAR-10\\ ($\epsilon=16/255$)}} & ResNet50 & Source & 99.9\tiny $\pm 0.024$ & 90.32\tiny $\pm 0.192$ & 99.22\tiny $\pm 0.017$ \\ \cline{2-6}
    & DenseNet-121 & Clean & 89.36\tiny $\pm 0.348$ & 76.5\tiny $\pm 0.198$ & \textbf{95.51}\tiny $\pm 0.082$\\ \cline{2-6}
    & VGG19-BN & Clean & 82.90\tiny $\pm 0.206$ & 79.17\tiny $\pm 0.306$ & \textbf{90.8}\tiny $\pm 0.086$ \\ \cline{2-6}
    & Inception-V3 & Clean & 86.61\tiny $\pm 0.181$ & 79.7\tiny $\pm 0.234$ & \textbf{93.49}\tiny $\pm 0.044$ \\ \cline{2-6}
    & WideResNet-28 & Robust & 19.9\tiny $\pm 0.098$ & 22.42\tiny $\pm 0.148$ & \textbf{23.90}\tiny $\pm 0.035$\\ \cline{2-6}
    & {PreActResNet-18}$^1$ & Robust & 19.98\tiny $\pm 0.069$ & 21.67\tiny $\pm 0.103$  & \textbf{23.53}\tiny $\pm 0.041$ \\ \cline{2-6}
    & {PreActResNet-18}$^2$ & Robust & 23.23\tiny $\pm 0.124$ & 24.49\tiny $\pm 0.071$ & \textbf{26.03}\tiny $\pm 0.039$ \\ 
    \Xhline{1.5pt}
    \end{tabular} 
    \caption{Comparison of transfer attack success rates in 10-step attack for PGD, PGD with raw update, and RGD. 
    In ImageNet, PreActResNet-18 model is from \citet{wongfast}, WideResNet-50 model is from \citet{salman2020adversarially}, ResNet-50 model is from \citet{robustness},
    In CIFAR-10, WideResNet-28 model is from \citet{dingmma}, {PreActResNet-18}$^1$ model is from \citet{wongfast},
    {PreActResNet-18}$^2$ model is from \citet{andriushchenko2020understanding}.}
    \label{tab:transfer_study}
\end{table*}

\begin{figure*}[ht]
    \centering
    \subfigure{\includegraphics[width=0.32\linewidth]{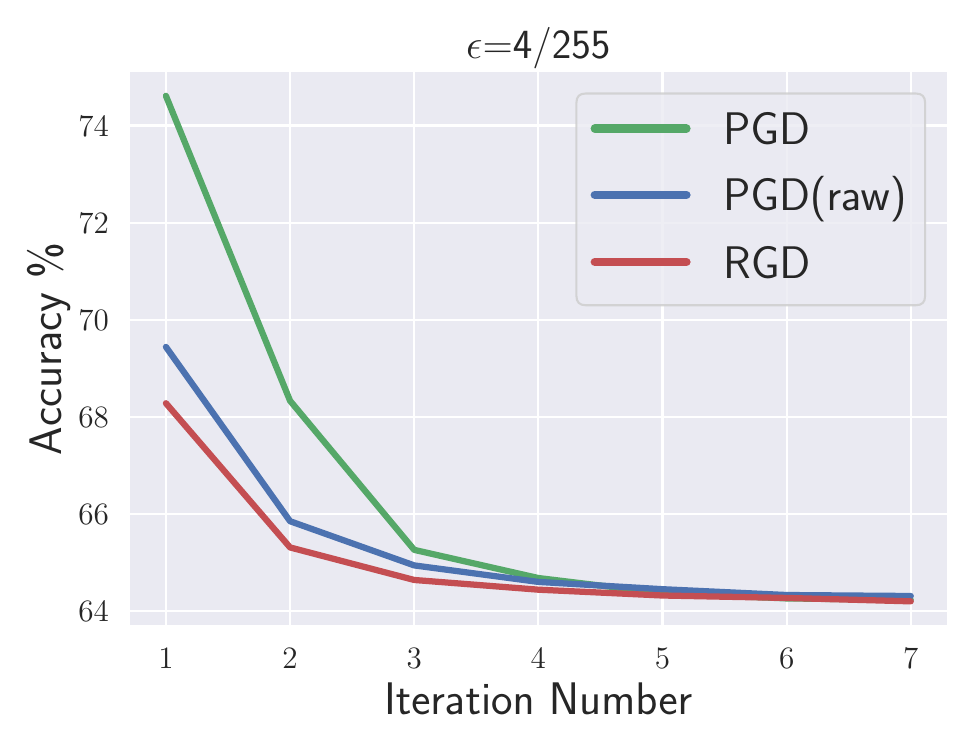}}
    \subfigure{\includegraphics[width=0.32\linewidth]{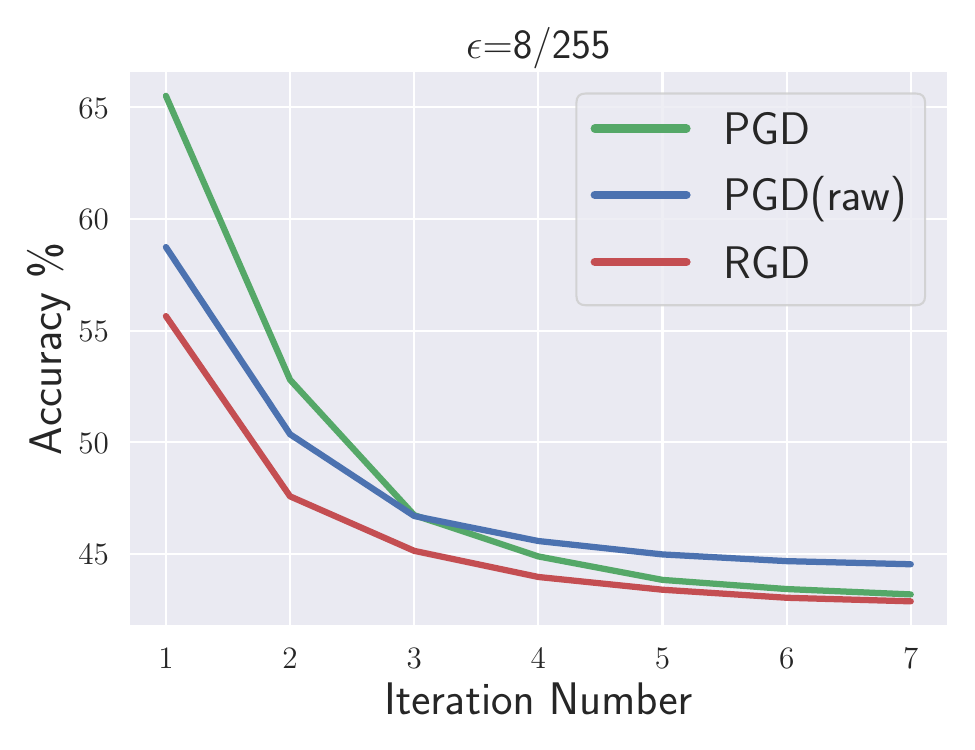}}
    \subfigure{\includegraphics[width=0.32\linewidth]{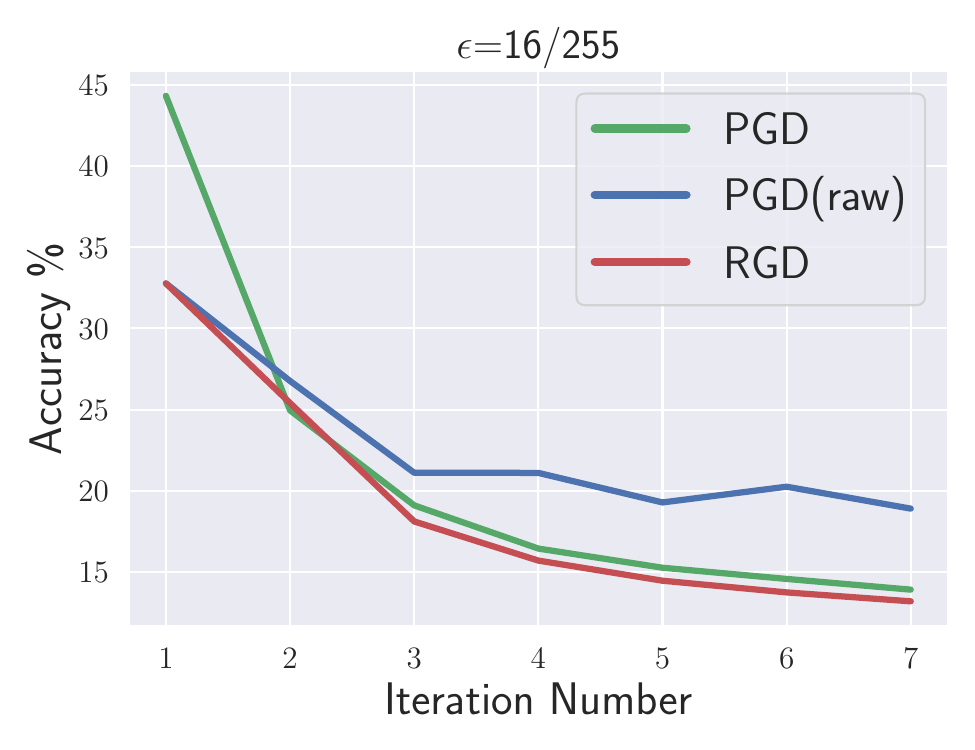}}
    \caption{Comparison of robust accuracy of PGD, PGD with raw update and proposed RGD with different $\epsilon$ sizes.}
    \label{fig:diff_eps}
\end{figure*}

\begin{figure*}[ht]
    \centering
    \subfigure{\includegraphics[width=0.32\linewidth]{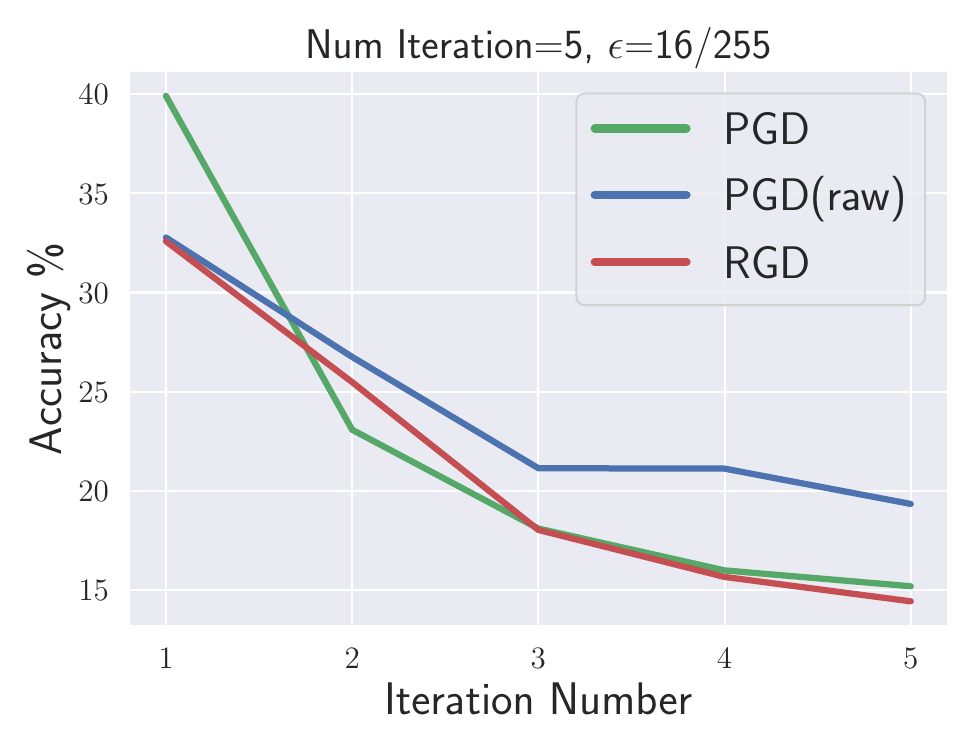}}
    \subfigure{\includegraphics[width=0.32\linewidth]{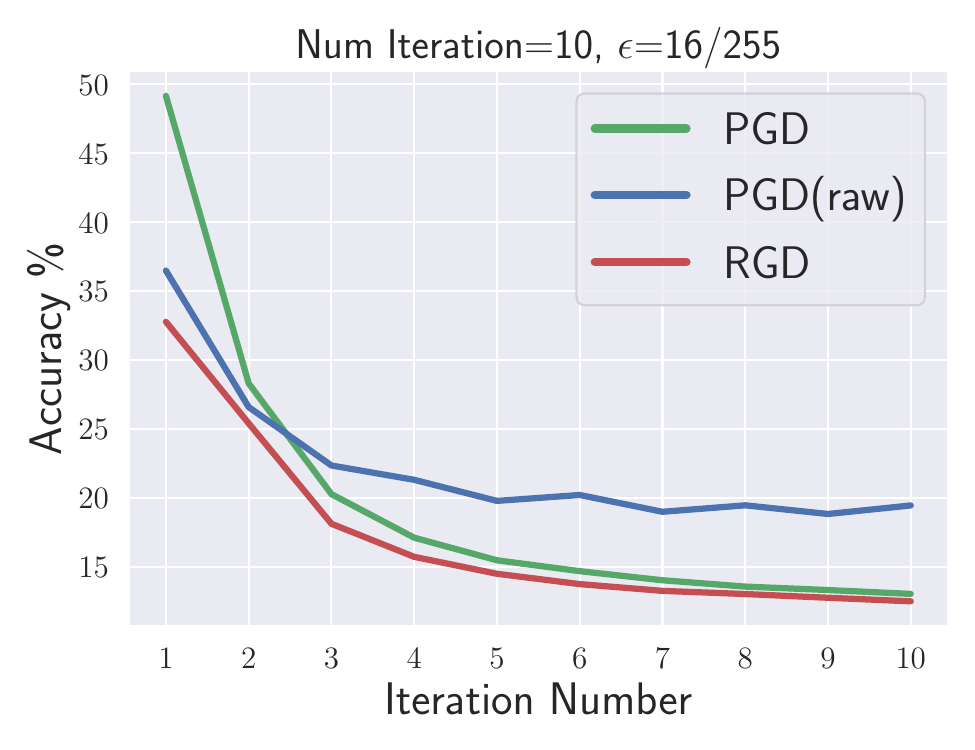}}
    \subfigure{\includegraphics[width=0.32\linewidth]{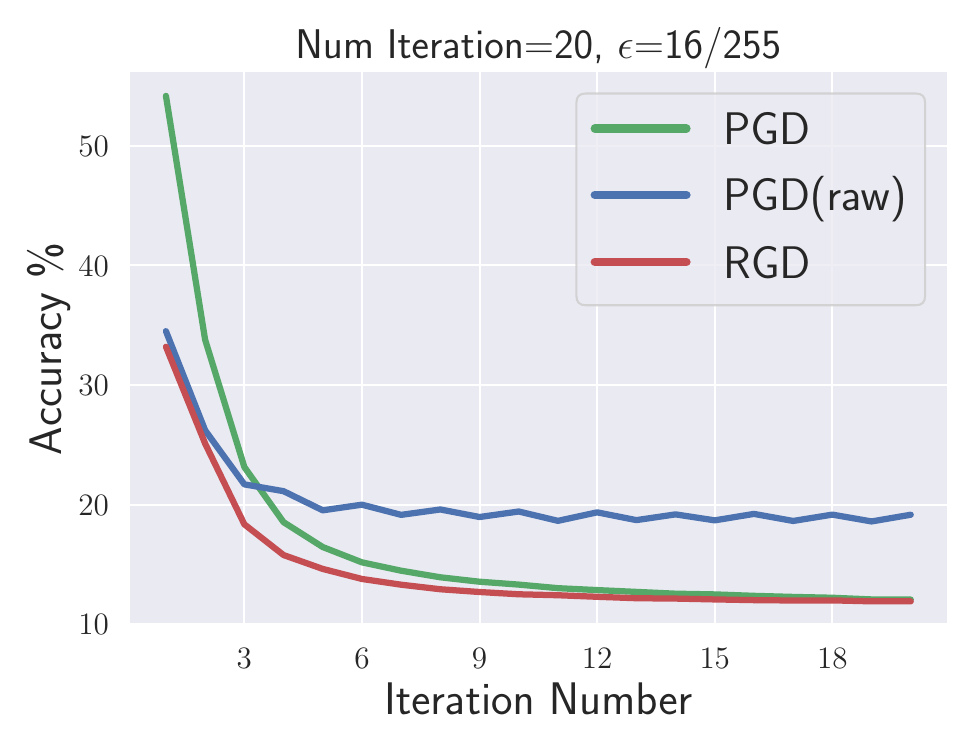}}
    \caption{Comparison of robust accuracy of PGD, PGD with raw update and RGD with different update steps when attacking robust ResNet18 model in CIFAR10.}
    \label{fig:diff_steps}
\end{figure*}

\subsection{Transfer Attack Study}

In this section, we examine and compare the transferability of RGD with PGD (sign/raw). We begin by generating adversarial data through a 10-step attack on the clean ResNet50 model~\citep{he2016deep}. These adversarial samples are then transferred to attack clean~\citep{simonyan2014very, huang2017densely, szegedy2015going} and robust~\citep{croce2021robustbench} target models. The ImageNet validation set and CIFAR-10 testing set are used as the data for the attacks, following the settings in~\citet{zhao2022towards}. The detailed results of the attack success rates for the source and target models can be found in \Cref{tab:transfer_study}.

The findings demonstrate that RGD consistently achieves the highest target success rates, while maintaining similar source success rates compared to PGD. It consistently improves the success rate by at least 5\% when attacking most clean models and by around 3\% for robust models with larger boundary ($\epsilon=16/255$). Furthermore, PGD (raw) outperforms signed PGD when attacking robust models, indicating that raw updates enhance the transferability to robust models. Our proposed RGD method further improves both robust and clean transferability compared to PGD (raw).

\section{Ablation and Visualization}

\subsection{Adversarial Perturbation Level Study}

In this section, we conduct a comprehensive comparison under different levels of adversarial perturbation. Specifically, we consider three different values of $\epsilon$ ($\epsilon=4/255, 8/255, 16/255$) for generating adversarial examples, and evaluate the robust accuracy of PGD with sign or raw gradient updates, as well as RGD, when attacking the robust ResNet18 model using the same settings as in \Cref{subsec:boundary_study}. We perform a grid search to determine the step sizes for each algorithm, and the results are presented in \Cref{fig:diff_eps}, where a lower robust accuracy indicates a stronger attack.

Our result reveals the following observations: In the context of small $\epsilon$ ($\epsilon=4/255$), all algorithms converge to a similar performance point, suggesting comparable attack effectiveness. However, for larger size ($\epsilon=16/255$), PGD with raw gradient update performs relatively poorly, while RGD outperforms the other algorithms with noticeable improvements. Therefore, when a larger perturbation is allowed, RGD is the preferred choice for achieving stronger attack performance.

Furthermore, the raw update-based algorithms, namely PGD (raw) and RGD, exhibit a lower robust accuracy in the early steps. It is important to mention that we utilize zero initial for PGD (raw) and RGD, while a stronger uniform random initialization for PGD. This suggests that the raw update enables the generation of high-quality adversarial examples in the early steps.

\subsection{Adversarial Update Step Study}

\label{subsec:update_steps}
In this section, we compare the performance of PGD with sign/raw updates and RGD at different update steps. Specifically, we evaluate the algorithms at steps 5, 10 and 20. The experimental setup follows \Cref{subsec:boundary_study} where step sizes are carefully fine-tuned. We fix the adversarial size $\epsilon$ at $16/255$, and the results are presented in \Cref{fig:diff_steps}.

Our findings reveal that in the early stages, RGD achieves a lower robust accuracy compared to PGD, which can be attributed to its generation of stronger adversarial examples through a genuine adversarial distribution. However, as the number of training steps increases, the performance gap between RGD and PGD diminishes. This can be explained by the stable perturbation change introduced by the sign function in PGD, allowing it to achieve comparable performance to RGD in the later steps. In summary, RGD is a preferable choice for scenarios requiring a few-step update.

Additionally, it is evident that PGD with raw update remains in a suboptimal performance region and shows limited improvement in the later steps. This observation further highlights the advantage of the non-clipping design in RGD, which facilitates the generation of stronger adversarial examples.

\section{Conclusion}

This work provides a theoretical analysis of how update procedures impact adversarial performance and offers an empirical explanation for why the sign operation is preferred in PGD. Additionally, we introduce the hidden unclipped perturbation and propose a novel attack algorithm called RGD. This algorithm transforms the constrained optimization problem into an unconstrained optimization problem. Extensive experiments have been conducted to demonstrate the superiority of proposed algorithm in practical scenarios.

\bibliographystyle{abbrvnat}
\bibliography{adv}

\begin{thebibliography}{44}
\providecommand{\natexlab}[1]{#1}
\providecommand{\url}[1]{\texttt{#1}}
\expandafter\ifx\csname urlstyle\endcsname\relax
  \providecommand{\doi}[1]{doi: #1}\else
  \providecommand{\doi}{doi: \begingroup \urlstyle{rm}\Url}\fi

\bibitem[Addepalli et~al.(2022)Addepalli, Jain, et~al.]{addepalli2022efficient}
S.~Addepalli, S.~Jain, et~al.
\newblock Efficient and effective augmentation strategy for adversarial
  training.
\newblock \emph{Advances in Neural Information Processing Systems (NeurIPS)},
  35:\penalty0 1488--1501, 2022.

\bibitem[Agarwal et~al.(2020)Agarwal, Singh, and Vatsa]{agarwal2020role}
A.~Agarwal, R.~Singh, and M.~Vatsa.
\newblock The role of'sign'and'direction'of gradient on the performance of cnn.
\newblock In \emph{Proceedings of the IEEE/CVF Conference on Computer Vision
  and Pattern Recognition Workshops}, pages 646--647, 2020.

\bibitem[Al-Dujaili and O'Reilly(2020)]{al2020sign}
A.~Al-Dujaili and U.-M. O'Reilly.
\newblock Sign bits are all you need for black-box attacks.
\newblock In \emph{International Conference on Learning Representations}, 2020.

\bibitem[Andriushchenko and Flammarion(2020)]{andriushchenko2020understanding}
M.~Andriushchenko and N.~Flammarion.
\newblock Understanding and improving fast adversarial training.
\newblock \emph{Advances in Neural Information Processing Systems (NeurIPS)},
  33:\penalty0 16048--16059, 2020.

\bibitem[Andriushchenko et~al.(2020)Andriushchenko, Croce, Flammarion, and
  Hein]{andriushchenko2020square}
M.~Andriushchenko, F.~Croce, N.~Flammarion, and M.~Hein.
\newblock Square attack: a query-efficient black-box adversarial attack via
  random search.
\newblock In \emph{European Conference on Computer Vision (ECCV)}, pages
  484--501. Springer, 2020.

\bibitem[Arora et~al.(2019)Arora, Du, Hu, Li, and Wang]{arora2019fine}
S.~Arora, S.~Du, W.~Hu, Z.~Li, and R.~Wang.
\newblock Fine-grained analysis of optimization and generalization for
  overparameterized two-layer neural networks.
\newblock In \emph{International Conference on Machine Learning (ICML)}, pages
  322--332, 2019.

\bibitem[Carlini and Wagner(2017)]{carlini2017towards}
N.~Carlini and D.~Wagner.
\newblock Towards evaluating the robustness of neural networks.
\newblock In \emph{2017 ieee symposium on security and privacy (sp)}, pages
  39--57. Ieee, 2017.

\bibitem[Chen and Gu(2020)]{chen2020rays}
J.~Chen and Q.~Gu.
\newblock Rays: A ray searching method for hard-label adversarial attack.
\newblock In \emph{Proceedings of the 26th ACM SIGKDD International Conference
  on Knowledge Discovery \& Data Mining (KDD)}, pages 1739--1747, 2020.

\bibitem[Chen et~al.(2017)Chen, Zhang, Sharma, Yi, and Hsieh]{chen2017zoo}
P.-Y. Chen, H.~Zhang, Y.~Sharma, J.~Yi, and C.-J. Hsieh.
\newblock Zoo: Zeroth order optimization based black-box attacks to deep neural
  networks without training substitute models.
\newblock In \emph{Proceedings of the 10th ACM workshop on artificial
  intelligence and security}, pages 15--26, 2017.

\bibitem[Croce and Hein(2020)]{croce2020reliable}
F.~Croce and M.~Hein.
\newblock Reliable evaluation of adversarial robustness with an ensemble of
  diverse parameter-free attacks.
\newblock In \emph{International Conference on Machine Learning (ICML)}, pages
  2206--2216. PMLR, 2020.

\bibitem[Croce et~al.(2021)Croce, Andriushchenko, Sehwag, Debenedetti,
  Flammarion, Chiang, Mittal, and Hein]{croce2021robustbench}
F.~Croce, M.~Andriushchenko, V.~Sehwag, E.~Debenedetti, N.~Flammarion,
  M.~Chiang, P.~Mittal, and M.~Hein.
\newblock Robustbench: a standardized adversarial robustness benchmark.
\newblock In \emph{Thirty-fifth Conference on Neural Information Processing
  Systems Datasets and Benchmarks Track}, 2021.
\newblock URL \url{https://openreview.net/forum?id=SSKZPJCt7B}.

\bibitem[Debenedetti et~al.(2022)Debenedetti, Sehwag, and
  Mittal]{debenedetti2022light}
E.~Debenedetti, V.~Sehwag, and P.~Mittal.
\newblock A light recipe to train robust vision transformers.
\newblock \emph{arXiv preprint arXiv:2209.07399}, 2022.

\bibitem[Ding et~al.(2019)Ding, Wang, and Jin]{ding2019advertorch}
G.~W. Ding, L.~Wang, and X.~Jin.
\newblock {AdverTorch} v0.1: An adversarial robustness toolbox based on
  pytorch.
\newblock \emph{arXiv preprint arXiv:1902.07623}, 2019.

\bibitem[Ding et~al.(2020)Ding, Sharma, Lui, and Huang]{dingmma}
G.~W. Ding, Y.~Sharma, K.~Y.~C. Lui, and R.~Huang.
\newblock Mma training: Direct input space margin maximization through
  adversarial training.
\newblock In \emph{International Conference on Learning Representations
  (ICLR)}, 2020.

\bibitem[Dong et~al.(2018)Dong, Liao, Pang, Su, Zhu, Hu, and
  Li]{dong2018boosting}
Y.~Dong, F.~Liao, T.~Pang, H.~Su, J.~Zhu, X.~Hu, and J.~Li.
\newblock Boosting adversarial attacks with momentum.
\newblock In \emph{Proceedings of the IEEE conference on computer vision and
  pattern recognition}, pages 9185--9193, 2018.

\bibitem[Du et~al.(2019)Du, Zhai, Poczos, and Singh]{dugradient}
S.~S. Du, X.~Zhai, B.~Poczos, and A.~Singh.
\newblock Gradient descent provably optimizes over-parameterized neural
  networks.
\newblock In \emph{International Conference on Learning Representations
  (ICLR)}, 2019.

\bibitem[Engstrom et~al.(2019)Engstrom, Ilyas, Salman, Santurkar, and
  Tsipras]{robustness}
L.~Engstrom, A.~Ilyas, H.~Salman, S.~Santurkar, and D.~Tsipras.
\newblock Robustness (python library), 2019.
\newblock URL \url{https://github.com/MadryLab/robustness}.

\bibitem[Goodfellow et~al.(2015)Goodfellow, Shlens, and
  Szegedy]{goodfellow2014explaining}
I.~J. Goodfellow, J.~Shlens, and C.~Szegedy.
\newblock Explaining and harnessing adversarial examples.
\newblock In \emph{International Conference on Learning Representations
  (ICLR)}, 2015.

\bibitem[He et~al.(2016)He, Zhang, Ren, and Sun]{he2016deep}
K.~He, X.~Zhang, S.~Ren, and J.~Sun.
\newblock Deep residual learning for image recognition.
\newblock In \emph{Proceedings of the IEEE conference on computer vision and
  pattern recognition (CVPR)}, pages 770--778, 2016.

\bibitem[Hochreiter and Schmidhuber(1997)]{hochreiter1997long}
S.~Hochreiter and J.~Schmidhuber.
\newblock Long short-term memory.
\newblock \emph{Neural computation}, 9\penalty0 (8):\penalty0 1735--1780, 1997.

\bibitem[Huang et~al.(2017)Huang, Liu, Van Der~Maaten, and
  Weinberger]{huang2017densely}
G.~Huang, Z.~Liu, L.~Van Der~Maaten, and K.~Q. Weinberger.
\newblock Densely connected convolutional networks.
\newblock In \emph{Proceedings of the IEEE conference on computer vision and
  pattern recognition (CVPR)}, pages 4700--4708, 2017.

\bibitem[Huang et~al.(2021)Huang, Wang, Erfani, Gu, Bailey, and
  Ma]{huang2021exploring}
H.~Huang, Y.~Wang, S.~Erfani, Q.~Gu, J.~Bailey, and X.~Ma.
\newblock Exploring architectural ingredients of adversarially robust deep
  neural networks.
\newblock \emph{Advances in Neural Information Processing Systems (NeurIPS)},
  34:\penalty0 5545--5559, 2021.

\bibitem[Ilyas et~al.(2018)Ilyas, Engstrom, Athalye, and Lin]{ilyas2018black}
A.~Ilyas, L.~Engstrom, A.~Athalye, and J.~Lin.
\newblock Black-box adversarial attacks with limited queries and information.
\newblock In \emph{International conference on machine learning}, pages
  2137--2146. PMLR, 2018.

\bibitem[Jia et~al.(2022)Jia, Zhang, Wu, Ma, Wang, and Cao]{jia2022adversarial}
X.~Jia, Y.~Zhang, B.~Wu, K.~Ma, J.~Wang, and X.~Cao.
\newblock Las-at: adversarial training with learnable attack strategy.
\newblock In \emph{Proceedings of the IEEE/CVF Conference on Computer Vision
  and Pattern Recognition (CVPR)}, pages 13398--13408, 2022.

\bibitem[Krizhevsky et~al.(2012)Krizhevsky, Sutskever, and
  Hinton]{NIPS2012_c399862d}
A.~Krizhevsky, I.~Sutskever, and G.~E. Hinton.
\newblock Imagenet classification with deep convolutional neural networks.
\newblock In F.~Pereira, C.~Burges, L.~Bottou, and K.~Weinberger, editors,
  \emph{Advances in Neural Information Processing Systems (NeurIPS)},
  volume~25. Curran Associates, Inc., 2012.

\bibitem[Kurakin et~al.(2018)Kurakin, Goodfellow, and
  Bengio]{kurakin2018adversarial}
A.~Kurakin, I.~J. Goodfellow, and S.~Bengio.
\newblock Adversarial examples in the physical world.
\newblock In \emph{Artificial intelligence safety and security}, pages 99--112.
  Chapman and Hall/CRC, 2018.

\bibitem[Liu et~al.(2019)Liu, Chen, Chen, and Hong]{liu2019signsgd}
S.~Liu, P.-Y. Chen, X.~Chen, and M.~Hong.
\newblock signsgd via zeroth-order oracle.
\newblock In \emph{International conference on learning representations}.
  International Conference on Learning Representations (ICLR), 2019.

\bibitem[Madry et~al.(2018)Madry, Makelov, Schmidt, Tsipras, and
  Vladu]{madry2018towards}
A.~Madry, A.~Makelov, L.~Schmidt, D.~Tsipras, and A.~Vladu.
\newblock Towards deep learning models resistant to adversarial attacks.
\newblock In \emph{International Conference on Learning Representations
  (ICLR)}, 2018.

\bibitem[Moosavi-Dezfooli et~al.(2016)Moosavi-Dezfooli, Fawzi, and
  Frossard]{moosavi2016deepfool}
S.-M. Moosavi-Dezfooli, A.~Fawzi, and P.~Frossard.
\newblock Deepfool: a simple and accurate method to fool deep neural networks.
\newblock In \emph{Proceedings of the IEEE conference on computer vision and
  pattern recognition (CVPR)}, pages 2574--2582, 2016.

\bibitem[Nair and Hinton(2010)]{nair2010rectified}
V.~Nair and G.~E. Hinton.
\newblock Rectified linear units improve restricted boltzmann machines.
\newblock In \emph{Proceedings of the 27th international conference on machine
  learning (ICML)}, 2010.

\bibitem[Rebuffi et~al.(2021)Rebuffi, Gowal, Calian, Stimberg, Wiles, and
  Mann]{rebuffi2021fixing}
S.-A. Rebuffi, S.~Gowal, D.~A. Calian, F.~Stimberg, O.~Wiles, and T.~Mann.
\newblock Fixing data augmentation to improve adversarial robustness.
\newblock \emph{arXiv preprint arXiv:2103.01946}, 2021.

\bibitem[Rice et~al.(2020)Rice, Wong, and Kolter]{rice2020overfitting}
L.~Rice, E.~Wong, and Z.~Kolter.
\newblock Overfitting in adversarially robust deep learning.
\newblock In \emph{International Conference on Machine Learning (ICML)}, pages
  8093--8104, 2020.

\bibitem[Salman et~al.(2020)Salman, Ilyas, Engstrom, Kapoor, and
  Madry]{salman2020adversarially}
H.~Salman, A.~Ilyas, L.~Engstrom, A.~Kapoor, and A.~Madry.
\newblock Do adversarially robust imagenet models transfer better?
\newblock \emph{Advances in Neural Information Processing Systems (NeurIPS)},
  33:\penalty0 3533--3545, 2020.

\bibitem[Shaham et~al.(2015)Shaham, Yamada, and
  Negahban]{shaham2015understanding}
U.~Shaham, Y.~Yamada, and S.~Negahban.
\newblock Understanding adversarial training: Increasing local stability of
  neural nets through robust optimization.
\newblock \emph{arXiv preprint arXiv:1511.05432}, 2015.

\bibitem[Simonyan and Zisserman(2015)]{simonyan2014very}
K.~Simonyan and A.~Zisserman.
\newblock Very deep convolutional networks for large-scale image recognition.
\newblock In \emph{International Conference on Learning Representations
  (ICLR)}, 2015.

\bibitem[Szegedy et~al.(2013)Szegedy, Zaremba, Sutskever, Bruna, Erhan,
  Goodfellow, and Fergus]{szegedy2013intriguing}
C.~Szegedy, W.~Zaremba, I.~Sutskever, J.~Bruna, D.~Erhan, I.~Goodfellow, and
  R.~Fergus.
\newblock Intriguing properties of neural networks.
\newblock \emph{arXiv preprint arXiv:1312.6199}, 2013.

\bibitem[Szegedy et~al.(2015)Szegedy, Liu, Jia, Sermanet, Reed, Anguelov,
  Erhan, Vanhoucke, and Rabinovich]{szegedy2015going}
C.~Szegedy, W.~Liu, Y.~Jia, P.~Sermanet, S.~Reed, D.~Anguelov, D.~Erhan,
  V.~Vanhoucke, and A.~Rabinovich.
\newblock Going deeper with convolutions.
\newblock In \emph{Proceedings of the IEEE conference on computer vision and
  pattern recognition (CVPR)}, pages 1--9, 2015.

\bibitem[Wang et~al.(2023)Wang, Pang, Du, Lin, Liu, and Yan]{wang2023better}
Z.~Wang, T.~Pang, C.~Du, M.~Lin, W.~Liu, and S.~Yan.
\newblock Better diffusion models further improve adversarial training.
\newblock \emph{arXiv preprint arXiv:2302.04638}, 2023.

\bibitem[Wong et~al.(2020)Wong, Rice, and Kolter]{wongfast}
E.~Wong, L.~Rice, and J.~Z. Kolter.
\newblock Fast is better than free: Revisiting adversarial training.
\newblock In \emph{International Conference on Learning Representations
  (ICLR)}, 2020.

\bibitem[Wu et~al.(2020)Wu, Xia, and Wang]{wu2020adversarial}
D.~Wu, S.-T. Xia, and Y.~Wang.
\newblock Adversarial weight perturbation helps robust generalization.
\newblock \emph{Advances in Neural Information Processing Systems (NeurIPS)},
  33:\penalty0 2958--2969, 2020.

\bibitem[Zhang et~al.(2020)Zhang, Zhu, Niu, Han, Sugiyama, and
  Kankanhalli]{zhang2020geometry}
J.~Zhang, J.~Zhu, G.~Niu, B.~Han, M.~Sugiyama, and M.~Kankanhalli.
\newblock Geometry-aware instance-reweighted adversarial training.
\newblock \emph{arXiv preprint arXiv:2010.01736}, 2020.

\bibitem[Zhang et~al.(2021)Zhang, Gong, Liu, Niu, Tian, Han, Sch{\"o}lkopf, and
  Zhang]{zhang2021causaladv}
Y.~Zhang, M.~Gong, T.~Liu, G.~Niu, X.~Tian, B.~Han, B.~Sch{\"o}lkopf, and
  K.~Zhang.
\newblock Causaladv: Adversarial robustness through the lens of causality.
\newblock \emph{arXiv preprint arXiv:2106.06196}, 2021.

\bibitem[Zhang et~al.(2022)Zhang, Zhang, Khanduri, Hong, Chang, and
  Liu]{zhang2022revisiting}
Y.~Zhang, G.~Zhang, P.~Khanduri, M.~Hong, S.~Chang, and S.~Liu.
\newblock Revisiting and advancing fast adversarial training through the lens
  of bi-level optimization.
\newblock In \emph{International Conference on Machine Learning (ICML)}, pages
  26693--26712. PMLR, 2022.

\bibitem[Zhao et~al.(2022)Zhao, Zhang, Li, Sicre, Amsaleg, and
  Backes]{zhao2022towards}
Z.~Zhao, H.~Zhang, R.~Li, R.~Sicre, L.~Amsaleg, and M.~Backes.
\newblock Towards good practices in evaluating transfer adversarial attacks.
\newblock \emph{arXiv preprint arXiv:2211.09565}, 2022.

\end{thebibliography}



\clearpage
\appendix

\onecolumn

\noindent{\Large{\bf Supplementary Materials}}

\section{Comparison with AutoAttack}
\label{subsec:autoattack}

AutoAttack~\citep{croce2020reliable} is an ensemble adversarial algorithm known for its success in adversarial attacks. One of its components, APGD\textsubscript{CE}, is a variant of the PGD algorithm. The results from \Cref{subsec:boundary_study} and \ref{subsec:update_steps} demonstrate that RGD is capable of learning the genuine adversarial distribution, resulting in stronger attack in the early steps. In contrast, PGD benefits from the larger perturbation changes introduced by the sign function, which leads to great performance in the later steps. Therefore, we integrate RGD into the 100-step APGD\textsubscript{CE} algorithm by replacing the first two APGD updates with RGD updates. Specifically, we implement APGD\textsubscript{CE}+RGD without restarting, and RGD is initialized with zero values. The step size $\alpha$ for RGD is carefully fine-tuned based on the first two step performance. 
The selection of two steps was motivated by its ability to yield the most significant enhancements in most scenarios. A comprehensive comparison of the improvements observed in the initial two steps is available in \Cref{supp:aa_extra}.
Following the same experimental setup in \citet{croce2020reliable}, we present comparison between APGD\textsubscript{CE} and APGD\textsubscript{CE}+RGD for their final robust accuracy in \Cref{tab:autoattack}.

\begin{table}[h]
	\centering
	\begin{tabular}{c|c|c|c}
		\Xhline{1.5pt}
		Dataset & Method & APGD\textsubscript{CE}  & APGD\textsubscript{CE}+RGD \\
		\hline
		\multirow{3}{*}{\makecell{CIFAR-10\\ ($\epsilon=16/255$)}} & WRN-34\citep{huang2021exploring} & 23.48\tiny $\pm 0.036$ & \textbf{23.35}\tiny $\pm 0.028$ \\ \cline{2-4}
		& WRN-28\citep{wu2020adversarial} & 28.0\tiny $\pm 0.061$ & \textbf{27.81}\tiny $\pm 0.012$ \\ \cline{2-4}
		& PreRN-18\citep{wongfast} & 9.63\tiny $\pm 0.037$ & \textbf{9.58}\tiny $\pm 0.01$ \\ 
		\hline
		\hline
		\multirow{4}{*}{\makecell{CIFAR-100\\ ($\epsilon=16/255$)}} & WRN-28\citep{wang2023better} & 19.43\tiny $\pm 0.082$ & \textbf{19.31}\tiny $\pm 0.005$ \\ \cline{2-4}
		& WRN-70\citep{rebuffi2021fixing} & 16.88\tiny $\pm 0.039$ & \textbf{16.55}\tiny $\pm 0.002$ \\ \cline{2-4}
		& XCiT\citep{debenedetti2022light} & 13.77\tiny $\pm 0.082$ & \textbf{13.71}\tiny $\pm 0.004$ \\ \cline{2-4}
		& WRN-34\citep{jia2022adversarial} & 11.56\tiny $\pm 0.036$ & \textbf{11.28}\tiny $\pm 0.016$ \\  
		\hline
		\hline
		\multirow{4}{*}{\makecell{ImageNet\\($\epsilon=4/255$)}} & RN-18\citep{salman2020adversarially} & 29.32\tiny $\pm 0.033$ & \textbf{29.26}\tiny $\pm 0.020$ \\ \cline{2-4}
		& XCiT\citep{debenedetti2022light} & 42.73\tiny $\pm 0.010$  & \textbf{42.73}\tiny $\pm 0.010$ \\ \cline{2-4}
		& PreRN-18\citep{wongfast} & 27.18\tiny $\pm 0.023$ & \textbf{27.07}\tiny $\pm 0.035$ \\ \cline{2-4}
		& WRN-50\citep{salman2020adversarially} & 40.86\tiny $\pm 0.029$ & \textbf{40.76}\tiny $\pm 0.032$\\ \cline{2-4}
		\Xhline{1.5pt}
	\end{tabular}
	\caption{Comparison of robust accuracy in 100-step attack between APGD\textsubscript{CE} (AutoAttack) and APGD\textsubscript{CE}+RGD. The methods are abbreviated as in \Cref{tab:pgd_attack}.}
	\label{tab:autoattack}
\end{table}

Our results demonstrate that despite the initial disadvantage of zero initialization, RGD achieves a lower robust accuracy compared to APGD in the first two steps as shown in \Cref{supp:aa_extra}, highlighting its superiority. Furthermore, RGD maintains this advantage in the final stage across most scenarios as shown in \Cref{tab:autoattack}. It is worth noting that such considerable performance improvement is achieved by only modifying the first two steps out of the 100 updates.

\section{Comparison of Initials}
\label{app:initial}

For initialization, we choose methods based on their suitability: PGD favors random initialization, while RGD and PGD (raw) lean towards zero initialization. To illustrate, we present the robust accuracy outcomes when attacking the robust ResNet18 model using various initial values:

\begin{table}[h]
    \centering
    \begin{tabular}{c|c|c}
        \Xhline{1.5pt}
         Method & Random Initial & Zero Initial  \\
        \hline
          PGD & 43.2\% & 43.27\%  \\
        \hline
        PGD (raw)  & 44.65\% & 44.49\%   \\
        \hline
         RGD  & \textbf{42.91\%} & \textbf{42.88\%} \\
        \hline
        \Xhline{1.5pt}
    \end{tabular}
    \caption{Robust accuracy comparison of PGD, PGD (raw) and RGD witin random/zero initial for 7-step attack.}
    \label{tab:initial}
\end{table}

From \Cref{tab:initial}, it is evident that PGD benefits from random initialization, whereas PGD (raw) and RGD perform better with zero initialization. Moreover, RGD exhibits superior performance in both contexts.

\section{Pixel-wise Experiments}
\label{app:pixel-wise}

To better illustrate that PGD enjoys larger perturbation change per step compared with RGD, we calculate the average pixel-wise perturbation change when attacking the robust ResNet18 model. The results are shown below:

\begin{table}[h]
    \centering
    \begin{tabular}{c|c|c|c|c|c|c|c|c}
        \Xhline{1.5pt}
        & Algorithm & 1 & 2 & 3 & 4 & 5 & 6 & 7 \\
        \hline
        \multirow{3}{*}{\makecell{Perturbation change}} & PGD & 0.0167 & 0.0119 & 0.0077 & 0.005 & 0.0041 & 0.0035 & 0.0032 \\
        \cline{2-9}
        &PGD (raw)  & 0.0116 & 0.0082 & 0.006 & 0.0054 & 0.0049 & 0.0048 & 0.0048  \\
        \cline{2-9}
        & RGD  & 0.0146 & 0.0084 & 0.0048 & 0.003 & 0.0022 & 0.0019 & 0.0017\\
        \hline
        \Xhline{1.5pt}
    \end{tabular}
    \caption{Comparison of update algorithms for average perturbation change in different steps.}
    \label{tab:pixel-wise}
\end{table}

From \Cref{tab:pixel-wise}, it is evident that RGD undergoes smaller perturbation shifts than PGD, leading to a decreased boundary ratio as depicted in \Cref{tab:pertu_illu}. Despite these minor perturbation variations, RGD, benefiting from genuine adversarial perturbations, surpasses PGD in performance. However, the edge of this improvement narrows with increasing iterations.

In the case of PGD (raw), while it might show pronounced perturbation changes in later stages, its adversarial loss remains suboptimal, and it lacks consistent convergence stability. As a result, when compared to both PGD and RGD, performance of PGD (raw) is notably inferior.

\section{AutoAttack Extra Experimental Results}
\label{supp:aa_extra}

\begin{table*}[ht]
    \centering
    \begin{tabular}{c|c|c|c}
     \Xhline{1.5pt}
        Dataset & Method & APGD\textsubscript{CE}  & APGD\textsubscript{CE}+RGD \\
        \hline
        \multirow{3}{*}{\makecell{CIFAR10\\ ($\epsilon=16/255$)}} & WRN-34\citep{huang2021exploring} & 90.01$\rightarrow$47.73 & 91.23$\rightarrow$45.79 \\ \cline{2-4}
         & WRN-28\citep{wu2020adversarial} & 86.46$\rightarrow$41.71 & 88.25$\rightarrow$39.3 \\ \cline{2-4}
         & PreRN-18\citep{wongfast} & 81.85$\rightarrow$24.75 & 83.34$\rightarrow$21.72 \\ 
        \hline
        \hline
        \multirow{4}{*}{\makecell{CIFAR100\\ ($\epsilon=16/255$)}} & WRN-28\citep{wang2023better} & 70.72$\rightarrow$25.6 & 72.58$\rightarrow$24.17 \\ \cline{2-4}
        & WRN-70\citep{rebuffi2021fixing} & 61.73$\rightarrow$22.75 & 63.56$\rightarrow$21.44 \\ \cline{2-4}
        & XCiT\citep{debenedetti2022light} & 65.64$\rightarrow$19.88 & 67.34$\rightarrow$19.16 \\ \cline{2-4}
        & WRN-34\citep{jia2022adversarial} & 65.11$\rightarrow$21.44 & 67.31$\rightarrow$20.33 \\  
        \hline
        \hline
        \multirow{4}{*}{\makecell{ImageNet\\($\epsilon=4/255$)}} & RN-18\citep{salman2020adversarially} & 52.64$\rightarrow$31.28 & 52.88$\rightarrow$30.68 \\ \cline{2-4}
        & XCiT\citep{debenedetti2022light} & 72.04$\rightarrow$46.58  & 72.06$\rightarrow$45.72 \\ \cline{2-4}
        & PreRN-18\citep{wongfast} & 52.97$\rightarrow$30.94 & 53.46$\rightarrow$30.12 \\ \cline{2-4}
        & WRN-50\citep{salman2020adversarially} & 68.36$\rightarrow$44.96 & 68.62$\rightarrow$43.9\\ \cline{2-4}
        \Xhline{1.5pt}
    \end{tabular}
    \caption{Comparison of robust ccuracy between APGD (AutoAttack) and APGD+RGD. The table presents the change in accuracy during the first two steps. The methods are abbreviated as follows: XCiT-S12 is denoted as XCiT, WideResNet as WRN, ResNet as RN, and PreActResNet as PreRN.}
    \label{tab:supp_autoattack}
\end{table*}

In this section, we evaluate the improvement in the first two steps of accuracy for APGD and APGD+RGD. The results are shown in \Cref{tab:supp_autoattack}. We can observe that despite having a worse initialization (zero initialization), RGD achieves a lower robust accuracy after two steps compared to APGD, indicating its superiority. Furthermore, APGD+RGD maintains these advantages in the final accuracy, as demonstrated in \Cref{tab:autoattack}.

\section{Related Lemmas}
\label{app:lemmas}

\begin{lemma}
\label{lem:function_diff}
Considering activation function $h(x)$ as ReLU, i.e., $h(x)=\max(x,0)=\frac{|x|+x}{2}$, then we define $|\cdot| $ as element wise absolute operation, $w_{1+}=\frac{1}{2}(w_1+|w_1|), w_{1-}=\frac{1}{2}(w_1-|w_1|)$ and obtain
\begin{align*}
    w_1^T&h(w_2*(x+\delta_c^{t+1}))-w_1^Th(w_2*(x+\delta_c^{t})) \\
    \overset{(i)}=&\frac{1}{2} w_1^T(|w_2(x+\delta_c^{t+1})|-|w_2(x+\delta_c^{t})|) + \frac{1}{2}w_1^Tw_2(\delta_c^{t+1}-\delta_c^{t})\\
    \overset{(ii)}= & \frac{1}{2} w_{1+}^T (|w_2(x+\delta_c^{t+1})|-|w_2(x+\delta_c^{t})|) + \frac{1}{2} w_{1-}^T(|w_2(x+\delta_c^{t+1})|-|w_2(x+\delta_c^{t})|)+\frac{1}{2}w_1^Tw_2(\delta_c^{t+1}-\delta_c^{t}) \\
    \overset{(iii)}\leq & \frac{1}{2}w_{1+}^T(|w_2(x+\delta_c^{t+1})-w_2(x+\delta_c^{t})|)-\frac{1}{2}w_{1-}^T(|w_2(x+\delta_c^{t+1})-w_2(x+\delta_c^{t})|)+
    \frac{1}{2}w_1^Tw_2(\delta_c^{t+1}-\delta_c^{t}) \\
    = & \frac{1}{2}(w_{1+}^T-w_{1-}^T)(|w_2(x+\delta_c^{t+1})-w_2(x+\delta_c^{t})|) + \frac{1}{2}w_1^Tw_2(\delta_c^{t+1}-\delta_c^{t}) \\
     = & \frac{1}{2}(w_{1+}^T-w_{1-}^T)|w_2(\delta_c^{t+1}-\delta_c^{t})| + \frac{1}{2}w_1^Tw_2(\delta_c^{t+1}-\delta_c^{t}) \\
     \overset{(iv)}=& \frac{1}{2}|w_1^T||w_2(\delta_c^{t+1}-\delta_c^{t})| + \frac{1}{2}w_1^Tw_2(\delta_c^{t+1}-\delta_c^{t}),
\end{align*}
where $(i)$ follows because $h(x)=\frac{|x|+x}{2}$, $(ii)$ follows because $w_1^T=w_{1+}^T+w_{1-}^T$, $(iii)$ follows because all elements in $w_{1+}^T$ are non-negative, all elements in $w_{1-}^T$ are non-positive and $(iv)$ follows because $w_{1+}-w_{1-}=|w_1|$. Then, we take absolute operation for both sides and obtain:
\begin{align*}
    |w_1^T&h(w_2*(x+\delta_c^{t+1})) -w_1^Th(w_2*(x+\delta_c^{t}))| \leq |w_1^T||w_2(\delta_c^{t+1}-\delta_c^{t})|\leq |w_1^T||w_2||\delta_c^{t+1}-\delta_c^{t}|,
\end{align*}
where $|\cdot| $ is defined as element wise absolute operation.

\end{lemma}

\begin{lemma}
\label{lem:estimation_error}
Considering $g(x)=\frac{1}{2}[w_1^Th(w_2*x)-y^\ast(x)]^2$, then we obtain
\begin{align*}
    \bigg|\frac{1}{2}&w_1^Th(w_2*(x+\delta_c^{t+1}))+\frac{1}{2}w_1^Th(w_2*(x+\delta_c^{t+1}))-y^\ast(x)\bigg | \\
    \leq & \frac{1}{2} |w_1^Th(w_2*(x+\delta_c^{t+1}))-y^\ast(x)| + \frac{1}{2} |w_1^Th(w_2*(x+\delta_c^{t}))-y^\ast(x)| \\
    \overset{(i)}= & \frac{\sqrt{2}}{2}[\sqrt{g(x+\delta_c^{t+1})} + \sqrt{g(x+\delta_c^{t})}],
\end{align*}
where $(i)$ follows from $g(x)$ definition.
\end{lemma}

\section{Proof of Theorem \ref{thm:main}}
\label{app:theorem_proof}
Considering $g(x)=\frac{1}{2}[w_1^Th(w_2*x)-y^\ast(x)]^2$, activation function $h(x)$ as ReLU, we define $|\cdot| $ as element wise absolute operation and characterize the adversarial step gain $g(x+\delta_c^{t+1})-g(x+\delta_c^t)$ as follows:
\begin{align*}
    g(x&+\delta_c^{t+1})-g(x+\delta_c^t)\\
    \overset{(i)}=& \frac{1}{2}[w_1^Th(w_2*(x+\delta_c^{t+1}))-y^\ast(x)]^2 - \frac{1}{2}[w_1^Th(w_2*(x+\delta_c^{t}))-y^\ast(x)]^2 \\
    =&\frac{1}{2} [w_1^Th(w_2*(x+\delta_c^{t+1})) - w_1^Th(w_2*(x+\delta_c^{t}))][w_1^Th(w_2*(x+\delta_c^{t+1}))+w_1^Th(w_2*(x+\delta_c^{t}))] \\
    & - [w_1^Th(w_2*(x+\delta_c^{t+1}))-w_1^Th(w_2*(x+\delta_c^{t}))]y^\ast(x) \\
    =& [w_1^Th(w_2*(x+\delta_c^{t+1}))-w_1^Th(w_2*(x+\delta_c^{t}))]*[\frac{1}{2}w_1^Th(w_2*(x+\delta_c^{t+1}))+\frac{1}{2}w_1^Th(w_2*(x+\delta_c^{t+1}))-y^\ast(x)] \\
    \leq & |w_1^Th(w_2*(x+\delta_c^{t+1}))-w_1^Th(w_2*(x+\delta_c^{t}))|*\bigg|\frac{1}{2}w_1^Th(w_2*(x+\delta_c^{t+1}))+\frac{1}{2}w_1^Th(w_2*(x+\delta_c^{t+1}))-y^\ast(x)\bigg| \\
    \overset{(ii)}\leq & \frac{\sqrt{2}}{2}*|w_1^T||w_2||\delta_c^{t+1}-\delta_c^{t}|\left(\sqrt{g(x+\delta_c^{t+1})} + \sqrt{g(x+\delta_c^{t})}\right) ,
\end{align*}
where $(i)$ follows from $g(x+\delta_c^t)$ definition and $(ii)$ follows from \Cref{lem:function_diff} and \ref{lem:estimation_error}.



\end{document}